\documentclass[twoside,11pt]{article}

\usepackage{blindtext}

\usepackage{amsmath}
\usepackage{amsfonts}
\usepackage{bm}
\usepackage{bbm}
\usepackage{tikz}
\usepackage{adjustbox}
\usepackage{subfig}
\usepackage[page]{appendix}
\usepackage[preprint]{jmlr2e}

\hypersetup{ hidelinks }

\newcommand{\ie}{i.e., }
\newcommand{\eg}{e.g., }

\newcommand{\X}{\mathbf{X}}
\newcommand{\Y}{\mathbf{Y}}
\newcommand{\A}{\mathbf{A}}

\newcommand{\tildeX}{\tilde{\X}}
\newcommand{\tildeY}{\tilde{\Y}}

\usetikzlibrary{shapes,decorations,arrows,calc,arrows.meta,fit,positioning}
\tikzset{
    -Latex,auto,node distance =1 cm and 1 cm,semithick,
    state/.style ={ellipse, draw, minimum width = 0.7 cm},
    hiddenstate/.style ={ellipse, draw, dashed, minimum width = 0.7 cm},
    point/.style = {circle, draw, inner sep=0.04cm,fill,node contents={}},
    anticausal/.style={dashed},
    causal/.style={dashed},
    el/.style = {inner sep=2pt, align=left, sloped}
}

\usepackage{lastpage}
\jmlrheading{23}{2022}{1-\pageref{LastPage}}{1/21; Revised 5/22}{9/22}{21-0000}{Enik\H{o} Székely, Sebastian Sippel, Nicolai Meinshausen, Guillaume Obozinski and Reto Knutti}

\ShortHeadings{Robust D\&A under interventions}{Székely, Sippel, Meinshausen, Obozinski and Knutti}
\firstpageno{1}

\begin{document}

\title{Robust detection and attribution of climate change under interventions}

\author{\name Enik\H{o} Székely \email eniko.szekely@epfl.ch \\
       \addr Swiss Data Science Center\\
       EPFL and ETH Zurich\\
       1015 Lausanne, Switzerland
       \AND
       \name Sebastian Sippel \email sebastian.sippel@env.ethz.ch \\
       \addr Institute for Atmospheric and Climate Science\\
       \addr Seminar for Statistics\\
       ETH Zurich\\
       8092 Zurich, Switzerland
       \AND
       \name Nicolai Meinshausen \email meinshausen@stat.math.ethz.ch\\
       \addr Seminar for Statistics\\
       ETH Zurich\\
       8092 Zurich, Switzerland
       \AND
       \name Guillaume Obozinski \email guillaume.obozinski@epfl.ch\\
       \addr Swiss Data Science Center\\
       EPFL and ETH Zurich\\
       1015 Lausanne, Switzerland
       \AND
       \name Reto Knutti \email reto.knutti@env.ethz.ch\\
       \addr Institute for Atmospheric and Climate Science\\
       ETH Zurich\\
       8092 Zurich, Switzerland
       }
       
\editor{My editor}

\maketitle

\begin{abstract}%

Fingerprints are key tools in climate change detection and attribution (D\&A) that are used to determine whether changes in observations are different from internal climate variability (\textit{detection}), and whether observed changes can be assigned to specific external drivers (\textit{attribution}). We propose a direct D\&A  approach based on supervised learning to extract fingerprints that lead to robust predictions under relevant interventions on exogenous variables, \ie climate drivers other than the target. We employ \textit{anchor regression}, a distributionally-robust statistical learning method inspired by causal inference that extrapolates well to perturbed data under the interventions considered. The residuals from the prediction achieve either uncorrelatedness or mean independence with the exogenous variables, thus guaranteeing robustness. We define D\&A as a unified hypothesis testing framework that relies on the same statistical model but uses different targets and test statistics. In the experiments, we first show that the CO$_2$ forcing can be robustly predicted from temperature spatial patterns under strong interventions on the solar forcing. Second, we illustrate attribution to the greenhouse gases and aerosols while protecting against interventions on the aerosols and CO$_2$ forcing, respectively. Our study shows that incorporating robustness constraints against relevant interventions may significantly benefit detection and attribution of climate change.

\end{abstract}


\section{Introduction}

Detection and attribution (D\&A) is a methodology in climate science whose aim is to study the causal links between external drivers of the climate system and observed changes in climate variables \citep[IPCC 2021,][]{EyringEtAl2021}. It seeks to determine whether and to what extent external climate forcings, \eg anthropogenic or natural forcings, or other specific forcings such as greenhouse gases (GHGs) or aerosols, influence observed climate variables, \eg temperature, precipitation, or humidity. According to the IPCC standard definition \citep{EyringEtAl2021}, \textit{detection} is the process of determining whether there is an observed change in a climate variable that cannot be explained by internal variability alone but without providing a reason for that change, while \textit{attribution} is the process of providing reasons for the detected change. 
    
Detection and attribution methods often employ so-called ``fingerprints'' that encode the response of the climate system to external forcings. Fingerprints are typically obtained from physics-based model simulations in the form of spatial, vertical, or spatio-temporal patterns \citep{HegerlEtAl1996, SanterEtAl2013, SanterEtAl2018}. In traditional D\&A, observations and pre-industrial control simulations are projected onto these fingerprints to determine whether changes in observations are significantly different from pre-industrial control simulations, an indication of an external forcing (likely) influencing the climate system. Fingerprints may be optimized against the internal variability of the climate system through a technique called optimal fingerprinting \citep{Hasselmann1993, AllenTett1999, RibesEtAl2009}, leading to an increased signal-to-noise ratio under the assumption that climate models adequately simulate the spatio-temporal scales of internal variability retained in the detection and attribution analysis. 
    
Recent research in statistical and machine learning has shown that direct approaches to D\&A that rely on supervised learning models can successfully identify fingerprints that are representative of specific targets (metrics), \eg a given external forcing or an associated forced response. These fingerprints are extracted from (spatio-temporal) climate variables simulated by physical models where the target is known \citep{SzekelyEtAl2019, BarnesEtAl2019, BarnesEtAl2020, SippelEtAl2020, WillsEtAl2020}. The projection of new model simulations or observations onto the resulting fingerprints maps high-dimensional (spatial) patterns to low-dimensional (often one-dimensional) metrics that capture the contribution of the target in explaining the changes in the climate variables. The low-dimensional metric can either be used by itself as a test statistic or used to define other test statistics to assess detection or attribution, thus transforming a high-dimensional testing problem into a lower-dimensional one. Direct D\&A methods that rely on supervised learning use explicitly the information contained in the target variable to extract the fingerprint. In traditional D\&A, the main idea is instead to perform an unsupervised dimension reduction through empirical orthogonal function (EOF, or equivalently principal component analysis) analysis prior to regression \citep{AllenTett1999}, resulting in a set of a few most important modes of variability. The projection of the observations and forced simulations onto these key modes of variability leads to low-dimensional embeddings where detection and attribution are assessed. 
    
The challenge of using fingerprints extracted using supervised learning is that they are not \textit{per se} robust against changes, \ie so-called ``interventions'' or ``perturbations'', due to drivers of the climate other than the target of interest. For example, simultaneously occurring changes in other external forcings than the forcing considered in attribution, or changes in the distributional properties of internal variability (across models, or between models and observations) might impact the validity of the D\&A statements if such changes were to project onto the fingerprints. A known key limitation of traditional D\&A is that both detection and attribution require an adequate simulation of internal variability to avoid potentially overconfident D\&A statements \citep{BindoffEtAl2013, SanterEtAl2019, SippelEtAl2021}. This would happen, for example, if the magnitude of the projection of the main modes of internal variability, such as multidecadal variability or El-Ni\~{n}o Southern oscillation, onto the externally-driven D\&A fingerprints would be large. Hence, additional robustness constraints need to be incorporated in the D\&A methodology to guarantee the validity of the detection and attribution statements. In direct D\&A, detection is implicitly robust under the invariant internal variability assumption, \ie climate models simulate internal variability adequately, while attribution requires additional constrains to be incorporated in the statistical learning model. Constraints are required for attribution because the prediction of a given external forcing, \eg anthropogenic forcing, depends on changes in other external forcings, \eg solar or volcanic forcings, also called exogenous variables (more details in Sect.~\ref{sect:distribRobust}). To extract fingerprints that lead to robust predictions under possible changes and interventions, and consequently allow for a robust attribution, we need to impose constraints on the statistical model. We are interested here in robustness to changes in distribution, \ie distributional changes, that might occur between the training and testing phases, so-called \textit{distributional robustness}. If the prediction is done in a different setting or environment \citep{Buehlmann2020}, \eg under changed states of internal variability or different forcings, the goal is to ensure the robustness of the prediction for generalization to out-of-distribution data, \ie data from a different distribution than that observed during training. Robustness often comes at a certain cost in prediction accuracy, therefore a trade-off between robustness and accuracy needs to be made. The problem of distributional robustness is related to transfer learning in machine learning \citep{PanYang2010, RojasCarullaEtAl2018}, where one seeks some form of invariance of the machine learning model that can be transferred from one task or domain to another. Distributional robustness and transfer learning allow us to achieve good prediction results on unseen data from perturbed distributions or different tasks \citep{PetersEtAl2016}.  
    
In this paper, we outline a direct approach to D\&A as a hypothesis testing framework which relies on fingerprints that are robust against a type of interventions called shift interventions. We frame D\&A in a high-dimensional supervised linear setting where we learn a set of linear regression  coefficients, \ie the fingerprint, to predict a given forcing or a proxy of the forced response from a map of climate variables, \eg temperature, across an ensemble of climate model simulations. To protect against interventions unrelated to the target variable that can influence the prediction results for attribution, we use a distributionally-robust supervised learning model to extract the fingerprint. More specifically, we employ \textit{anchor regression} \citep{RothenhaeuslerEtAl2021}, a statistical learning method inspired by causal inference that incorporates robustness constraints to extrapolate well to perturbed data. Here, we are interested in distributional robustness against interventions on climate variables that are due to external drivers other than the target, so-called \textit{anchor} variables in anchor regression. Once the fingerprint is extracted, and identical to traditional D\&A, out-of-sample climate data, \eg data from a climate model not seen during fingerprint extraction, are projected onto the fingerprint to assess detection or attribution. Anchor regression achieves robustness by enforcing that the part of the residuals from the prediction explained by the anchor variables is as small as possible, \ie as much as possible of the dependence of the residuals on the anchor is removed (more details in Sect.~\ref{sect:anchorRegr}). We show in \ref{sect:ap_nonlinearAnchors} that incorporating nonlinear anchors into anchor regression leads to mean independence of the residuals with the anchor variables through connections to the correlation ratio \citep{Renyi1959}.

The paper is organized as follows. We first provide a motivating example to highlight the advantages of a robust framework for fingerprint extraction where the goal is to identify CO$_2$-induced climate change signals under strong (but a priori unknown) changes in the solar forcing in a climate model (Sect.~\ref{sect:motivExample}). The CO$_2$ forcing is accurately recovered from the spatial pattern of annual temperatures even under very strong interventions on the solar forcing. Section \ref{sect:DA} describes the detection and attribution methodology and the hypothesis testing framework, and Sect.~\ref{sect:distribRobust} outlines the notion of distributional robustness and robust D\&A fingerprints as well as the connection to distributional robustness in statistical and machine learning. Section \ref{sect:anchorRegr} describes how distributional robustness can be achieved through anchor regression and discusses how the constraints imposed by anchor regression on the residuals of the prediction achieve uncorrelatedness and mean independence with the exogenous variables.  The attribution examples in Sect.~\ref{sect:experiments} show that applying anchor regression in a ``model as truth'' setting across the CMIP6 (Climate Model Intercomparison Project, Version 6) multi-model archive allows us to predict with a high accuracy and robustness both the GHGs and aerosols forcing while protecting against changes in the aerosols and CO$_2$ forcing, respectively. The robustness achieved through anchor regression therefore improves the confidence in the attribution statements to the forcings considered across the CMIP6 multi-model archive.

\section{Motivating example: Predicting CO$_2$ forcing under strong solar interventions}
\label{sect:motivExample}

\begin{figure}
 \centerline{\includegraphics[width = 0.95\textwidth]{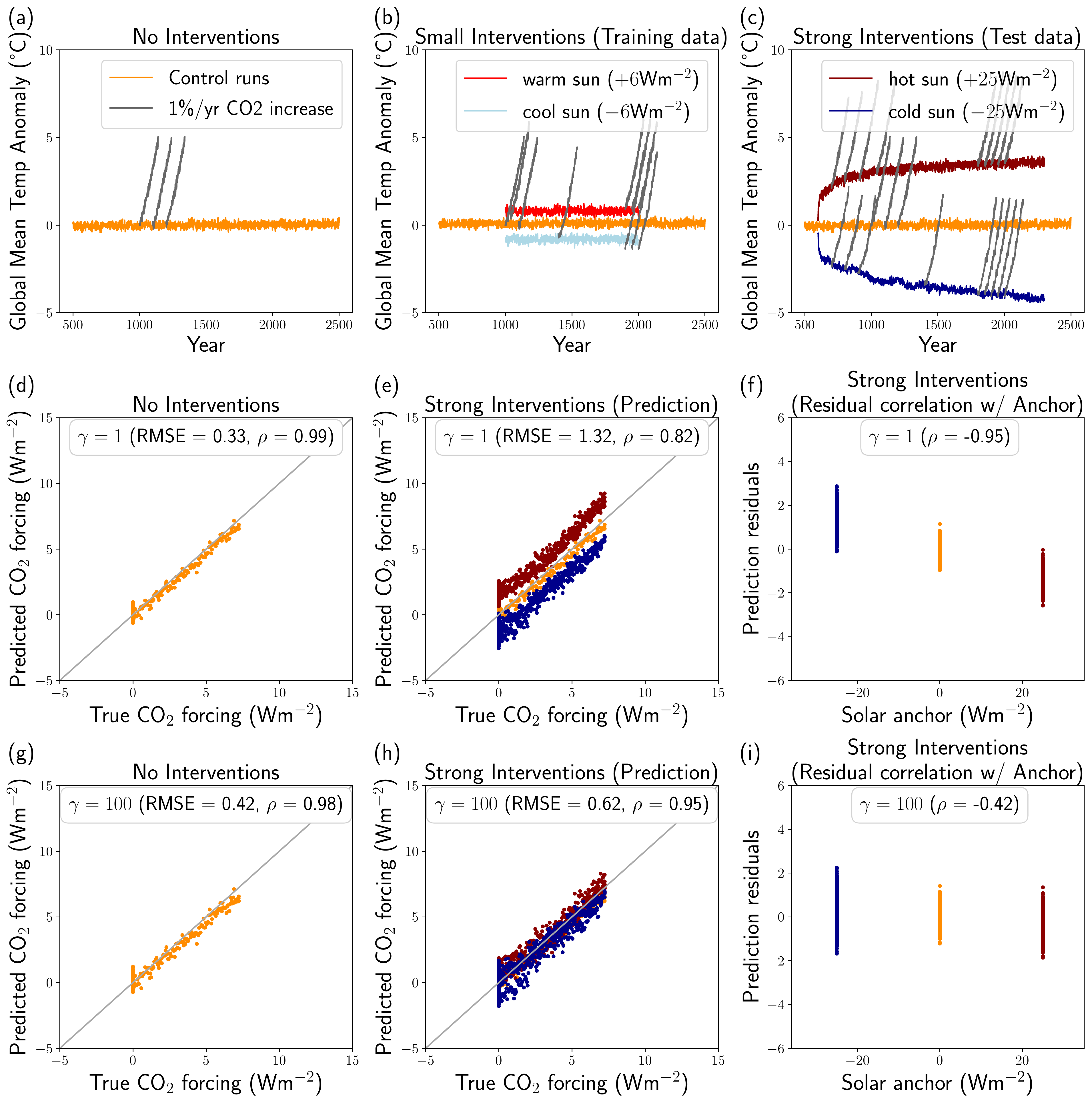}}
  \caption{(a) The pre-industrial control simulation (``normal'' solar state) is shown in orange with an approximate global mean surface temperature of about 13.8$^\circ$C. The data is centered prior to training. (b) Training dataset with only small solar variations ($\pm 6 \text{Wm}^{-2}$, imposed on the solar constant); red = ``warm sun'', light blue = ``cool sun''. 
   (c) Simulations with strong solar variations ($\pm 25 \text{Wm}^{-2}$, imposed on the solar constant), used as test dataset; dark red = ``hot sun'', dark blue = ``cold sun''. (a, b, c) grey = 1\% CO$_2$ concentration increase simulations that branch off each of the sun states. (d, g) Prediction for ``no interventions'' test data. (e, h) Prediction for ``strong interventions'' test data (where the solar variations in the test dataset exceed the solar variations in the training dataset by a factor of more than four). (f, i) Correlation of the residuals with the anchor variable, \ie solar forcing, for the ``strong interventions'' for $\gamma=1$ and $\gamma=100$.}\label{fig:solar}
\end{figure}

We start with a simple motivating example to illustrate the idea of D\&A fingerprints that are representative of a given forcing but also robust to the pattern imprint of potentially strong perturbations in other forcings \citep{SzekelyEtAl2019}, an important requirement for robust attribution. Here, we aim to predict the CO$_2$ forcing from spatial patterns of climate variables in idealized climate model simulations run by the Community Earth System Model \citep[Version 1.2.2;][]{StolpeEtAl2019}. These idealized simulations consist of three ``baseline'' simulations without any CO$_2$ forcing that differ in the magnitude of the solar forcing: 1) a ``hot'' (+25Wm$^{-2}$, imposed on the solar constant), 2) a ``cold'' (-25Wm$^{-2}$), and 3) a ``normal'' (0Wm$^{-2}$) solar state \citep{StolpeEtAl2019}, which is kept constant within each simulation. The ``normal'' solar state corresponds to a conventional pre-industrial control simulation in the absence of CO$_2$ forcing. In each simulation we introduce artificially-generated time segments of 140 years in which the CO$_2$ concentration is increased by 1\% per year until it has quadrupled compared to the pre-industrial era. Each of the 140 years segments branches off from the three solar baseline states described above as shown with grey lines in Figs.~\ref{fig:solar}a, b, c. An overview of the dataset, later used as test data, is shown in Fig.~\ref{fig:solar}c, and a detailed description is available in \citet{StolpeEtAl2019}.

The overall goal in this illustrative example is to find a fingerprint (in the form of a map of regression coefficients obtained from a regression model) that predicts the CO$_2$ forcing robustly from the temperature spatial pattern irrespective of whether the sun is in a ``hot'', ``cold'', or ``normal'' state. To show the advantages of robust fingerprints, \ie fingerprints that lead to robust predictions, we generate a training dataset (Fig.~\ref{fig:solar}b) by scaling the average solar-perturbed pattern such that the variations in the strength of the sun are only up to $\pm$6Wm$^{-2}$ (further referred to as the ``warm'' and ``cool'' solar states). This is only a small perturbation compared to the $\pm$25Wm$^{-2}$ variations in the test data, and would correspond to a solar radiative forcing $F_s$ of only about $\pm$1.05Wm$^{-2}$ when assuming a planetary albedo $\alpha_p$ of 0.3 ($F_s = \Delta S_0 \frac{1-\alpha_p}{4}$, with the solar constant $S_0$). The variations in the solar forcing in the training dataset are therefore ``underestimated'' by a factor of more than four compared to the test dataset. The fingerprints will be extracted using \textit{only} the small interventions ($\pm$6Wm$^{-2}$) from the training dataset, but would be required to achieve accurate predictions even under much stronger interventions in the test setting ($\pm$25Wm$^{-2}$). 

For prediction, we use anchor regression \citep{RothenhaeuslerEtAl2021} to obtain a model which is robust to variations in the intensity of the solar activity, in combination with a ridge penalty to avoid overfitting to the training data. We use as target variable for prediction the CO$_2$ forcing and as anchor variable the solar forcing (more details on anchor regression and anchor variables in Sect.~\ref{sect:anchorRegr}). Anchor regression has one hyperparameter $\gamma$ that controls the strength of the interventions on the anchor variable. In Fig.~\ref{fig:solar}, we illustrate results for two anchor regression hyperparameter values: 1) $\gamma = 1$, which is equivalent to ordinary least squares with a ridge penalty in our case, and hence minimizes the root mean squared error ($RMSE$) on the training dataset but protects only against very small interventions, and 2) $\gamma = 100$, which protects against relatively strong solar variations. Results are shown only for unseen test data points that have not been included in the training. The ridge regression hyperparameter is fixed here to $\lambda = 10^4$ (more details in Sect.~\ref{sect:anchorRegr}).

We first discuss results for the ``no interventions'' case where the CO$_2$ increases up to a quadrupling of pre-industrial concentrations while the sun remains in its ``normal'' unperturbed state throughout, \ie evaluating only against 1\% CO$_2$ per year increasing simulations that branch off of the orange line in Fig.~\ref{fig:solar}c. The prediction error for the CO$_2$ forcing is $RMSE =0.33$Wm$^{-2}$ for ridge regression (Fig.~\ref{fig:solar}d, $\gamma = 1$), and it increases to $RMSE =0.42$Wm$^{-2}$ for $\gamma=100$ (Fig.~\ref{fig:solar}g). The prediction error increases because anchor regression guards the predictions against potentially large interventions on the strength of the sun, but at a cost, \ie higher prediction error, in case those interventions do not occur at testing time. However, in both cases ($\gamma=1$ and $\gamma=100$) a reasonable prediction of the CO$_2$ forcing up to a quadrupling of pre-industrial CO$_2$ concentration  is achieved (\ie high correlation between observed and predicted values, Figs.~\ref{fig:solar}d, g).

Under ``strong interventions'', where the solar variations in the test dataset exceed the solar variations in the training dataset by a factor of more than four, the prediction error for ridge regression ($\gamma = 1$) increases from $RMSE = 0.33$Wm$^{-2}$ under the ``no interventions'' scenario (Fig.~\ref{fig:solar}d) to $RMSE = 1.32$Wm$^{-2}$ under strong interventions (Fig.~\ref{fig:solar}e). The large error occurs because the strong solar forcing (``hot sun'') causes a strong positive bias in the prediction of CO$_2$ forcing (Fig.~\ref{fig:solar}e). Conversely, the predictions of CO$_2$ against a ``cold sun'' background are biased downwards. Hence, while the strong interventions on the sun are unrelated to the CO$_2$ forcing itself, the former lead to an imprint on the pattern of surface temperatures that adversely affect and bias the CO$_2$ predictions. The impact of the changes in the solar state on the quality of the prediction can also be seen through the large negative correlation (Pearson linear correlation $\rho=-0.95$) of the residuals from the prediction (CO$_2^{\text{true}}$ - CO$_2^{\text{pred}}$) with the solar forcing (Fig.~\ref{fig:solar}f). In contrast, the prediction error for the ``anchor'' version of the fingerprint ($\gamma = 100$) increases only from $RMSE = 0.42$Wm$^{-2}$ in the unperturbed case (``no interventions'', Fig.~\ref{fig:solar}g) to $RMSE = 0.62$Wm$^{-2}$ in the perturbed case, \ie for the strong solar perturbations (Fig.~\ref{fig:solar}h). Moreover, systematic biases in the ``hot'' or ``cold'' sun scenarios are reduced strongly and almost fully disappear (Figs.~\ref{fig:solar}h, i). The constraint on solar forcing introduced by anchor regression  helps to extract a fingerprint that minimizes the adverse influence of solar variations on the prediction of CO$_2$. In other words, the anchor regression predictions are more (distributionally) robust with respect to solar variations compared to ordinary least squares. The correlation of the residuals with the anchor has been reduced from $\rho = -0.95$ (``no interventions'', Fig.~\ref{fig:solar}f) to $\rho = -0.42$ (``strong interventions'', Fig.~\ref{fig:solar}i). Overall, this methodology may allow us to avoid biases that could potentially affect D\&A statements in cases where a given real-world climate forcing would be underestimated in climate models and if variations in that forcing were to project onto fingerprints representative of other forcings. 

\section{Detection and attribution as hypothesis testing}
\label{sect:DA}

Traditional D\&A relies on regression models, typically linear approaches, to explain the observations as a function of model simulations driven with different external forcings. Due to the large (and increasing) spatial resolution of climate measurements, the regression cannot be performed directly on the full spatial data. Traditional D\&A therefore first projects the data into a lower-dimensional subspace of the most representative fingerprints (in terms of maximum variance) extracted using an EOF analysis (or principal component analysis, PCA), and the regression is performed in this subspace, of typically relatively small dimensionality \citep{AllenTett1999,AllenStott2003,RibesEtAl2009}. Another traditional detection approach extracts fingerprints of forced signals from transient simulations without optimization against internal variability characteristics (so-called ``non-optimal detection''), and subsequently assesses detection using a one-dimensional test statistic obtained by projecting the observations and pre-industrial control simulations onto those fingerprints \citep{HegerlEtAl1996, SanterEtAl2009,MarvelBonfils2013,SanterEtAl2018,SanterEtAl2019}. 

The direct approach presented here provides an alternative to the fingerprint extraction employed in traditional D\&A. Instead of performing an unsupervised dimension reduction, we use directly the information contained in the external radiative forcing or an associated target metric within climate model simulations to extract the fingerprint in a supervised way. In a high-dimensional linear regression setting, the fingerprint that most accurately captures the climate response to the target metric will be the map of regression coefficients. We note that our primary goal is not to predict the target variable, but rather to use it to define a test statistic based on which we formulate a unified framework for detection and attribution as hypothesis testing. The methodology for both detection and attribution relies on the same statistical model, but uses different target variables and test statistics. We discuss that while detection is implicitly robust under certain assumptions, robust attribution requires a constrained statistical model to account for potential changes in drivers of the climate other than the target, \eg other external forcings than the forcing of interest in attribution.

\subsection{Direct approach to detection and attribution}

Given a high-dimensional predictor variable $X \in \mathbb{R}^p$ we want to predict a target variable $Y \in \mathbb{R}$. The predictor $X$ is the (global or local) spatial map of a climate variable (\eg temperature, precipitation, humidity) at a given time $t$, and is of dimensionality $p$ given by the number of spatial grid cells. The target variable $Y$ can either be an external radiative forcing (anthropogenic, aerosols, natural, etc.) or a combination of radiative forcings, concentrations or emissions, or any other (global or local) target metric that measures the impact of the external forcings, such as the forced climate response. 

Let $f_{\beta}(X)$ be a function parameterized by $\beta$ that predicts $Y$ from $X$. The parameters $\beta$ are estimated by minimizing a given loss function $\ell  ( Y ,  f_{\beta}(X) ) $ over a sample drawn from the population distribution $(X, Y) \sim P$,
\begin{equation}
\label{eq:populationLoss}
\hat{\beta} = \operatorname*{argmin}_{\beta} \mathbb{E}_{(X, Y) \sim P} [\ell ( Y ,  f_{\beta}(X) ) ].
\end{equation}

The prediction model $f_{\beta}(\cdot)$ can be any regression model, \eg a linear \citep{SippelEtAl2020} or nonlinear (kernel) regression model, a random forest or a deep neural network \citep{BarnesEtAl2019}. The predicted target for a given $x$ is $\hat{Y}(x) = f_{\hat{\beta}}(x)$, where $\hat{Y}$ is the predicted target variable. One challenge with the estimator $\hat{\beta}$ as defined in \eqref{eq:populationLoss} is that it only optimizes over the observed population distribution $(X, Y)\ {\sim}\ P$ available at training time, and does not take into account possible changes in the distribution that could occur at testing time. These changes can in turn lead to poor prediction results for out-of-distribution data \citep{Meinshausen2018, Buehlmann2020} due to either large variance in the predictions or distributional bias. Bias can be particularly problematic in D\&A induced for example by reconstruction errors in the external forcings, \eg if the forced response due to an external driver were to be larger or smaller than in the model simulations, or under uncertain and potentially large internal variability \citep{SippelEtAl2021}. To address this problem, the optimization in \eqref{eq:populationLoss} can be extended to include an entire class of distributions instead of just the observed population distribution $P$, which will ensure robustness under distributional changes due to the external forcings. We note here that ``observed population distribution'' does not refer to the distribution of the observations as understood in climate science, rather it is simply the distribution of the data that is available at training time. 

The statistical model from \eqref{eq:populationLoss} is trained on climate data from model simulations. Let $x_*$ be an out-of-sample data point that comes either from climate models not seen during training (so-called ``model as truth'' experiment), from reanalysis, or directly from observations. The prediction $\hat{y}_* = \hat{Y}(x_*)$ for $x_*$ uses the regression coefficients $\hat{\beta}$ from \eqref{eq:populationLoss}, as
\begin{equation*}
\label{eq:prediction}
\hat{y}_* = f_{\hat{\beta}}(x_*),
\end{equation*}
\noindent where $x_*$ is a $p$-dimensional spatial map at a given time $t$, and $\hat{y}_*$ is the value of the predicted target variable at that time. We denote by $X_*$, $Y_*$, and $\hat{Y}_{*}$ the input, output, and predicted target variables for out-of-sample data, respectively. We focus in this paper on the ``model as truth'' experiment and leave the prediction of observations and reanalysis for future work. ``Model as truth'' experiments typically take a certain set of climate models as training or calibration data, and the calibrated method or statistical models are then tested on a yet unseen set of climate models. This is a standard approach in climate studies to test or benchmark analysis methods \citep{DeserEtAl2020}, also sometimes referred to as ``perfect model scenario'' or ``imperfect model''. In the following, we describe the hypothesis testing framework for detection and attribution and the test statistics defined based on the predicted target variable $\hat{Y}$. 

\subsection{Detection} 
\label{sect:detection}

We define the null hypothesis for detection to be the absence of an externally-forced climate change signal, and a given test statistic $T(\hat{Y})$ defined based on the predicted target variable $\hat{Y}$. We want to test against the null hypothesis that the test statistic is indistinguishable from internal variability, \ie no external forcing or causes are influencing the climate response and all changes can be explained solely from the variability internal to the climate system. Framed as a signal-to-noise problem, the signal is the externally-forced climate change and the noise is the internal variability. If we are able to reject the null hypothesis, then we say we have \textit{detected} an externally-forced climate change signal. Our definition of detection is in line with the definition from IPCC AR4 WG1 \citep[][Ch.~9]{HegerlEtAl2007} wherein ``detection is the process of demonstrating that climate has changed in some defined statistical sense, without providing a reason for that change''. Because estimates of internal variability cannot be derived from observations directly, we use climate model estimates of unforced variability, a standard practice in D\&A \citep{SanterEtAl1996, HegerlEtAl1996, SanterEtAl2019}. Hence, the detection hypothesis test makes the crucial assumption that model estimates of internal variability are reliable \citep{AllenTett1999,SanterEtAl2019}.   

\begin{figure}{
\subfloat[Detection]{\includegraphics[width = 0.45\textwidth]{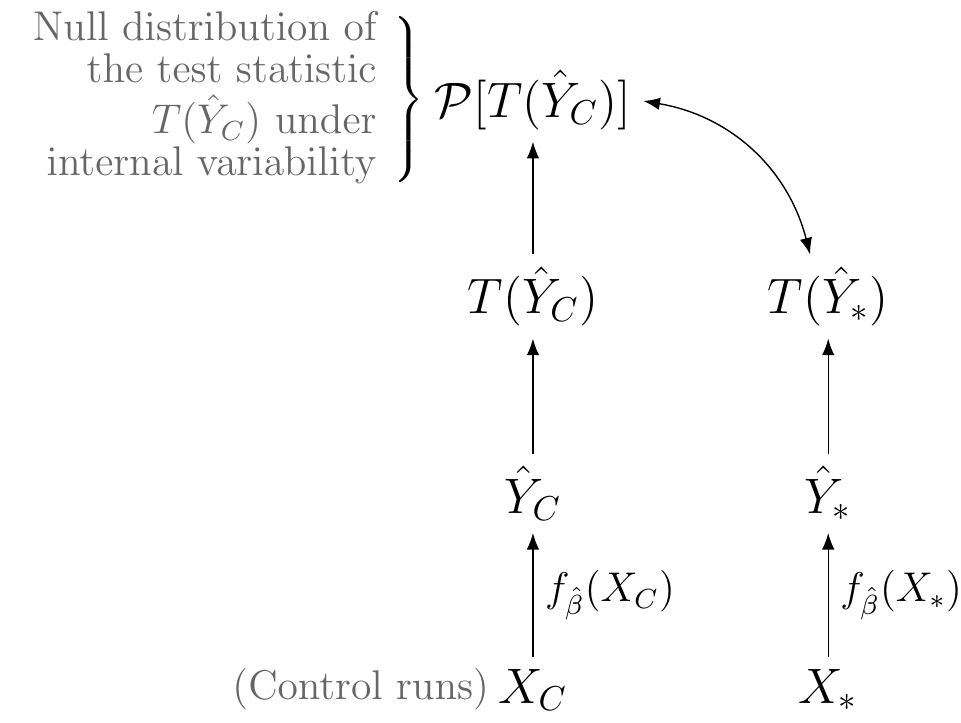}}
\hspace{0.5cm}
\subfloat[Attribution]{\includegraphics[width = 0.5\textwidth]{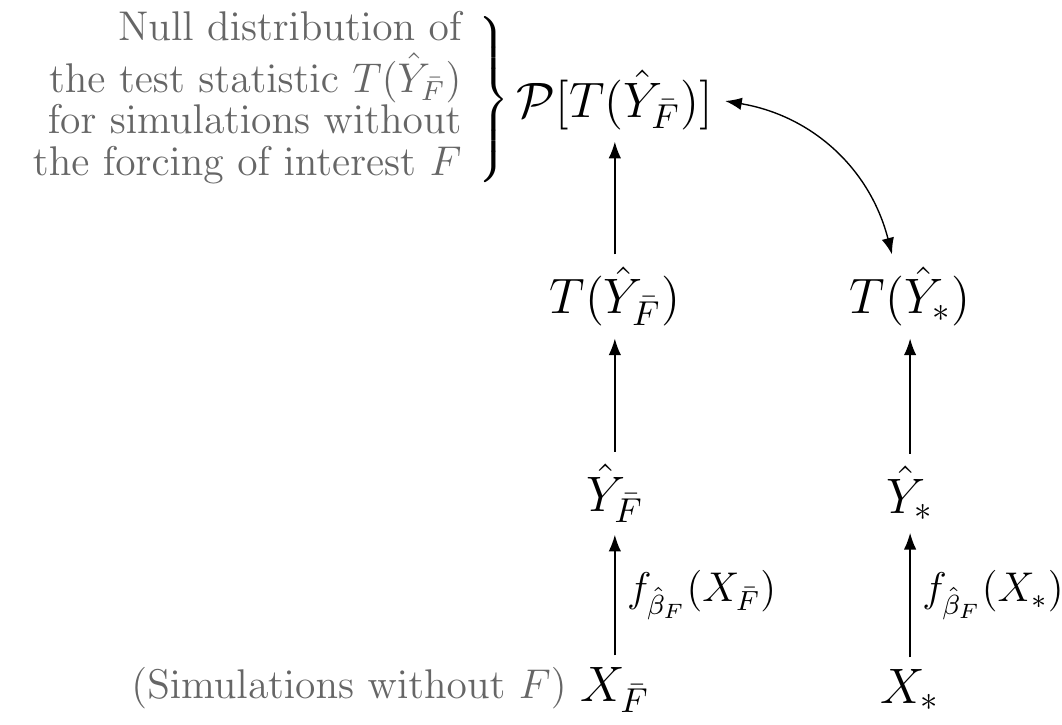}}
\caption{Direct detection and attribution: (a) \textbf{Detection}: First, we compute the predicted target variable $\hat{Y}_C$ from control runs $X_C$ (only internal variability) using the fingerprint $\hat{\beta}$ estimated using \eqref{eq:populationLoss}. The null distribution  $\mathcal{P}[T(\hat{Y}_C)]$ is the distribution of the test statistic $T(\hat{Y}_C)$ under the null hypothesis. The predicted target $\hat{Y}_*$ for out-of-sample data $X_*$ is computed using the same fingerprint $\hat{\beta}$. If the test statistic $T(\hat{Y}_*)$ falls in the critical region of the null distribution we say we \textit{detect} an externally-forced climate change signal. (b) \textbf{Attribution}: The methodology is similar to detection, but here the fingerprint $\hat{\beta}_F$ is estimated using a target variable $Y$ derived from the forcing of interest $F$ for which we want to test attribution. The null distribution is the distribution of the test statistic $T(\hat{Y}_{\bar{F}})$ from simulations $X_{\bar{F}}$ that do not contain the forcing $F$. If the test statistic $T(\hat{Y}_*)$ for out-of-sample data $X_*$ falls in the critical region of the null distribution, we say we \textit{attribute} (part of) the detected change to the forcing $F$. \label{fig:detection}}}
\end{figure}

We first estimate the regression coefficients $\hat{\beta}$ using both externally-forced and unforced (only internal variability) climate simulations using the statistical model in \eqref{eq:populationLoss}. Because detection aims to assess if there is an externally-forced climate change signal regardless of the cause or the source of external forcing, the statistical model is trained using a target metric $Y$ that captures the impact of the \textit{total} external forcing that generated $X$. We denote by $X_C$ the input variable for data from unforced simulations (control runs) and the corresponding predicted target variable by $\hat{Y}_C$. Detection consists in comparing the test statistic for out-of-sample data $T(\hat{Y}_*)$ from models not seen during training against the null distribution, \ie the distribution of the test statistic under the null hypothesis, given by $\mathcal{P}[T(\hat{Y}_C)]$ and estimated only from internal variability. If $T(\hat{Y}_*)$ falls in the critical region, \ie the rejection region of the null distribution (or equivalently, outside the confidence intervals), we reject the null hypothesis and say that we have detected an externally-forced climate change signal. Otherwise, if we fail to reject the null hypothesis, we cannot make any detection statements. We use a ``climate model split'' approach  for training and testing where only control runs $X_C$ from climate models not seen during training are used to construct the critical region at testing time. This approach, common in D\&A, helps to remove intra-model biases, \ie within the same model between forced and unforced runs, however inter-model biases due to climate model dependence might still remain. In the experimental results (Sect.~\ref{sect:experiments}) we use aggregating to remove as much as possible of the remaining inter-model biases by constructing multiple model-based subsamples and then aggregating the results. The detection procedure is summarized in Fig.~\ref{fig:detection}a.

\subsection{Attribution} 
\label{sect:attribution}

Attribution studies investigate whether one can attribute a detected climate change signal to any of the external forcings, \eg anthropogenic, aerosols, natural, or a combination of these forcings. Unequivocal attribution would require controlled experimentation with the climate system, \ie interventions on the climate system (see Sect.~\ref{sect:distribRobust} for a more detailed discussion on interventions), that is impossible to achieve in practice  \citep[IPCC AR4 WG1,][]{HegerlEtAl2007}. Therefore, we need to restrict our attention to a small number of candidate forcings, a standard practice in D\&A \citep[\eg][]{Hasselmann1993}. Let us assume for example that the forcing of interest $F$ for which we want to test attribution is the anthropogenic forcing (but any other radiative forcing or target metric could be used as an example). In this case, the null hypothesis would state that the change detected is \textit{not} due to the anthropogenic forcing. If the null hypothesis is rejected, then we say we \textit{attribute} (at least part of) the detected change to the anthropogenic forcing. According to IPCC AR4 WG1 \citep{HegerlEtAl2007}, ``attribution of causes of climate change is the process of establishing the most likely causes for the detected change with some defined level of confidence''. Because different forcings, \eg anthropogenic and solar, can have similar spatial climate responses, uniquely quantifying the relative contribution of each separate cause only from the spatial response is not always possible. Here, we address this issue by testing attribution to a forcing of interest whilst separating the components of the climate response that are common to other forcings. 

The methodology is similar to detection and relies on the same statistical model in \eqref{eq:populationLoss},  but while for detection the target variable $Y$ is a measure of the impact of the total external forcing, for attribution the target variable is derived only from the forcing of interest $F$. One challenge when testing attribution to a given forcing is that other additional forcings can have a simultaneous causal effect on the climate response $X$. Perturbations on these forcings can lead to distributional changes impacting the predictive performance of the statistical model in \eqref{eq:populationLoss}, and could consequently mislead the attribution process. By contrast with detection which is implicitly robust under certain assumptions on the internal variability, \ie the noise (see Sect.\ \ref{sect:distribRobust}), to achieve a robust attribution we need to protect against these distributional changes such that the estimated fingerprint $\hat{\beta}_F$ not only captures the climate response to the forcing $F$ but is also robust to changes in other forcings. This requires a distributionally-robust version of the model in \eqref{eq:populationLoss} that also works well under perturbations (see details in Sect.~\ref{sect:anchorRegr}). We will call the new attribution methodology ``$\gamma$-attribution'' with $\gamma$ indicating the degree or level of robustness. 

The test statistic under the null hypothesis is computed from the predicted forcing $\hat{Y}_{\bar{F}} = f_{\hat{\beta}_F}(X_{\bar{F}})$, where $X_{\bar{F}}$ are climate simulations that are not driven with the forcing of interest $F$, denoted by $\bar{F}$, such as control runs. We again make sure that the simulations $X_{\bar{F}}$ only come from climate models not seen during training using the ``climate model split'' approach for training and testing. If the test statistic for out-of-sample data $T(\hat{Y}_*)$ falls in the critical region of the null distribution $\mathcal{P}[T(\hat{Y}_{\bar{F}})]$, we reject the null hypothesis and say that we attribute (part of) the detected change to the external forcing $F$. Otherwise, if we fail to reject the null hypothesis, we cannot make any statements about attribution to the forcing $F$. The changes detected could however be attributed to other forcings, which can be tested by defining a new null hypothesis and repeating the attribution process for each forcing considered. Nonetheless, it is possible to detect a change and not be able to attribute it to any forcing in particular, but only to a combination of forcings. If the forcing of interest $F$ that we want to test attribution to is the total external forcing, then detection and attribution are equivalent (using only control runs for $X_{\bar{F}}$), that is, we attribute the detected change to the total external forcing. The procedure for attribution is summarized in Fig.~\ref{fig:detection}b.

\subsection{Test statistics}
\label{sect:test_statistics}

The hypothesis testing procedure in Fig.~\ref{fig:detection} can be applied either for each time step separately \citep{SippelEtAl2020}, \eg to compare a given day or year to internal variability, or for entire time series, \eg at the level of a climate model run or observational record. In each case, the test statistic $T(\hat{Y})$ needs to be defined accordingly. For individual time steps, the hypothesis testing is done separately for each data point $x_*$ using the value of the test statistic at the corresponding time $T(\hat{y}_*)$. Temporal information can be incorporated a posteriori by using for example $L$-length trends of the test statistic \citep{SanterEtAl2019} with $L$ number of time steps. For detection or attribution at the level of a model run or observational record, the test statistics are defined as functions over time series, \ie the time series of the predicted target for a time interval of interest. We note that the predicted target $\hat{Y}$ is estimated at each time step separately and relies only on the spatial information (or potentially, a vertical or multi-variable pattern) captured in the fingerprint, but does not take into account the temporal evolution of the forcings. As mentioned above, this gives us the possibility to define the test statistic either for single time steps or entire time series. In the following, we discuss one instance of the latter case for attribution using the rank correlation as test statistic, however one can use any other statistic for the hypothesis testing, \eg other types of correlation, the projection on the principal components, etc. 

To test attribution to a given forcing $F$ (the detection procedure is identical but uses the total external forcing for $F$ as in Fig.~\ref{fig:detection}a), we choose as test statistic the Spearman rank correlation $r_s$ between the forcing $F$ and the predicted target $\hat{Y}$,
\begin{equation*}
    T(\hat{Y}) = r_s(Y, \hat{Y}) = r_s(Y, X \hat{\beta}_F), \quad \text{with}\ Y = F,
\end{equation*}
where $\hat{\beta}_F$ is the fingerprint representative of forcing $F$ learnt during training. The Spearman rank correlation is a nonparametric measure that assesses the statistical dependence between rankings of two variables, and therefore reflects not only linear (as the Pearson linear correlation) but all monotonic relationships, including nonlinear ones. However, with either the linear correlation or the rank correlation, information on the strength of the trends is lost. In this case, one could use $L$-length trends as mentioned above for the test statistic \citep{SanterEtAl2019}. The rank correlation takes as input two time series, \ie the true forcing of interest $F$ and the predicted target $\hat{Y}$ for a given run of a given model and for a given time interval (in the experimental setting in Sect.~\ref{sect:experiments} we use the entire historical period for CMIP6 from 1850-2014), and outputs a scalar value that is compared against the null distribution under internal variability. The null distribution, and its corresponding critical region, are obtained using the control runs of all the test models except the model that we want to test attribution to using a so-called ``leave-one-model-out'' strategy, \ie we leave out all the runs (forced and unforced) from a given model to avoid any overlap between the estimation of the null distribution and the estimation of the statistic being tested, similar to a ``model as truth'' experiment. This strategy is then straightforward to apply to observations, if one considers the observations as the model left out. The critical region of the null distribution is built as detailed in Sect.~\ref{sect:experiments} and \ref{sect:HT_appendix}. The type I error and power for each model are then estimated using the critical region and an aggregation method called subagging \citep[subsample aggregating,][]{BuehlmannYu2002} that constructs subsamples of models without replacement and then averages the results. The type I error is the probability of rejecting the null hypothesis when the null is actually true, \ie to wrongly detect/attribute the changes to the forcing $F$ in the absence of forcing; and the power measures the probability of rejecting the null hypothesis under the alternative, \ie to correctly detect/attribute the changes to the forcing of interest. Ideally, the type I error should be as close as possible to zero and the power as close as possible to one. Details on the mathematical derivations of the hypothesis testing procedure are provided in \ref{sect:HT_appendix}. 

We presented here the general hypothesis testing framework for detection and attribution, and in the next two sections we discuss the requirements that we need to impose on the statistical model in \eqref{eq:populationLoss} to achieve a robust attribution.

\section{Distributional robustness and robust fingerprints}
\label{sect:distribRobust}

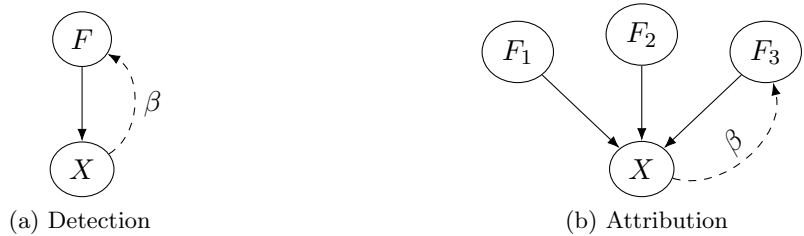
\begin{figure}
\begin{center}
\subfloat[Detection]{
\begin{adjustbox}{width=2.3cm}
\begin{tikzpicture}
	\node at (-1,0) {};
    \node[state] (x) at (0,0) {$X$};
    \node[state] (y) [above =of x] {$F$};

    \path (y) edge (x);

    \path[anticausal] (x) edge[bend right = 60] node[right] {$\beta$} (y);
\end{tikzpicture}
\end{adjustbox}}
\hspace{4cm}
\subfloat[Attribution]{\begin{tikzpicture}
    \node[state] (x) at (0,0) {$X$};
    \node[state] (y1) [above left =of x] {$F_1$};
    \node[state] (y2) [above =of x] {$F_2$};
    \node[state] (y3) [above right =of x] {$F_3$};

    \path (y1) edge (x);
    \path (y2) edge (x);
    \path (y3) edge (x);

    \path[anticausal] (x) edge[bend right = 60] node[el, above] {$\beta$} (y3);
\end{tikzpicture}
}
\caption{Causal diagrams for detection and attribution. Multiple external forcings $F_1, F_2,$ and $F_3$, \eg solar, volcanic, and anthropogenic forcing, have a direct causal effect (plain arrows) on the climate variable $X$, \eg temperature, precipitation. (a) \textbf{Detection}: The effect of the total external forcing $F=F_1+F_2+F_3$ on the climate variable $X$, and the prediction (dashed arrow) using the statistical model in \eqref{eq:populationLoss} for $Y = F$. (b) \textbf{Attribution}: Testing attribution to one forcing, \eg anthropogenic forcing $F_3$, requires being robust to changes due to other forcings, \eg solar $F_1$ and volcanic $F_2$ forcings, because of their direct effect on the climate variable $X$. The prediction (dashed arrow) for $Y = F_3$ uses the distributionally-robust estimator in \eqref{eq:distrRobust}.}
\label{fig:causal}
\end{center}
\end{figure}

Let us consider a high-dimensional climate variable $X$, such as temperature, precipitation, or humidity, that is potentially influenced by multiple external forcings. For illustration, we consider the causal diagram in Fig.~\ref{fig:causal}b with three external forcings, \eg solar $F_1$, volcanic $F_2$, and anthropogenic $F_3$, that have a direct causal effect on $X$, such that
$$X = g(F_1, F_2, F_3, \varepsilon),$$ 
where $\varepsilon$ is the internal variability or noise, and $g$ is a function of the different forcings and the noise. The diagram can be extended to include other forcings if necessary, or to split the anthropogenic forcing into its constituent parts, \eg GHGs, aerosols. If we want for example to test attribution to the anthropogenic forcing $Y = F_3$, we seek to learn a fingerprint $\hat{\beta} = \hat{\beta}_{F_3}$ that best explains the forcing of interest $F_3$ while also being robust to changes in $F_1$ and $F_2$ that directly influence $X$. 

The regression model in \eqref{eq:populationLoss} provides an estimator for $\beta$ that is optimal for the target distribution $P$, meaning that the estimator is a good predictor of the value of the forcing $Y$ for the given distribution. Under the assumption that the internal variability, \ie the noise distribution, is the same in the training and test data, such a statistical model is sufficient for detection. This is due to the fact that detection considers the total external forcing, \ie $F$ in Fig.~\ref{fig:causal}a, and does not assume the existence of any other unknown external causes or factors, meaning that there are no hidden variables in the system that could lead to changes in the distribution $P$. We additionally make the assumption that the physical mechanism captured by the conditional distribution of $X$ given the target $Y$, $p(X \mid Y)$, for $Y = F$, that generated both the training and test data is the same, \ie the climate models used for training are representative of the entire population of climate models, the mechanism is invariant in time, and the noise model is the same. If the internal variability (noise $\varepsilon$) changes, for example due to different low-frequency variabilities across climate models, the changes in the noise distribution could also be accounted for using the distributionally-robust approach described here to make detection of the forced warming robust to differences in internal variability \citep{SippelEtAl2021}. If the mechanism changes over time, as is the case for the aerosols spatial imprint for example, the regression model finds a combination of the changing patterns (see the experiments in Sect.\ \ref{sect:experiments}).

On the contrary, in the case of attribution to a forcing of interest, for example the anthropogenic forcing $F_3$ in Fig.~\ref{fig:causal}b, there are other \textit{known} factors in the system, such as the solar $F_1$ or volcanic $F_2$ forcings, that also have a direct effect on the climate response $X$. Despite their direct effect on $X$, they do not appear in the regression model from \eqref{eq:populationLoss}, and therefore any changes in $X$ occurring at testing time due to these external forcings cannot be accounted for and can bias the prediction results. A change in $F_1$ or $F_2$, \eg a stronger solar or volcanic forcing, leads to a change in $X$ and in the conditional distribution of $X$ given the target $Y$, $p(X \mid Y)$, for $Y = F_3$ (\eg as illustrated in Sect.~\ref{sect:motivExample} for the hypothetical case of a very strong or very weak solar forcing at testing time). Because the joint distribution $P = p(X,Y)$ from \eqref{eq:populationLoss} can be written as $p(X,Y) = p(X \mid Y) p(Y)$, a change in the conditional distribution $p(X \mid Y)$ \citep[also called conditional shift, ][]{ZhangEtAl2013} leads to a change in the joint distribution $p(X,Y)$ from the training to the testing stages even as the distribution of the target variable $p(Y)$ remains the same.  As a consequence, the anthropogenic forcing fingerprint $\hat{\beta}_{F_3}$ learnt during training using the distribution $P$ from \eqref{eq:populationLoss} might not lead to good predictions in a climate with a stronger solar or volcanic forcing, which could potentially mislead the attribution process. Our goal is to guarantee that the null distribution of the test statistic $\mathcal{P}[T(\hat{Y}_{\bar{F}})]$, $F = F_3$, from Sect.~\ref{sect:attribution} does not change (or changes as little as possible) under distributional changes (changes in $P$) due to these external forcings, meaning that $\hat{Y}_{\bar{F}} = f_{\hat{\beta}_F}(X_{\bar{F}})$ is a distributionally-robust estimate of $F$ as discussed below. Analogous to detection, we make the assumption that the physical mechanism given by $p(X \mid F_1, F_2, F_3)$ remains the same, \ie the climate models used for training are representative of the entire population of climate models, the mechanism is invariant in time, and the noise model is the same.  

A robust attribution can be achieved by reformulating the optimization problem in \eqref{eq:populationLoss} from a distributional robustness perspective \citep{Meinshausen2018, Buehlmann2020} to ensure that the prediction model $f_{\beta}(\cdot)$ captures the specific effect of the targeted forcing regardless of perturbations on the other external forcings, also called exogenous variables. The distributionally-robust form of the estimator in \eqref{eq:populationLoss} is given by 
\begin{equation}
\label{eq:distrRobust}
\hat{\beta} = \operatorname*{argmin}_{\beta} \sup_{Q \in \mathcal{Q}} \mathbb{E}_{(X,Y) \sim Q} [\ell ( Y ,  f_{\beta}(X) ) ],
\end{equation}
which optimizes not only for the target distribution $P$, but for the worst case distribution in a larger class of distributions $\mathcal{Q}$ generated by interventions. Distributional robustness is formulated here as a worst-case scenario, and thus ensures that the fingerprint $\hat{\beta}$, though conservative, leads to predictions that are robust under distributional changes. The fingerprint captures features that are shared across the class of distributions $\mathcal{Q}$, \ie invariant features, and we therefore call $\hat{\beta}$ a \textit{robust} or \textit{invariant} fingerprint. Robust predictions in the presence of external changes are necessary to guarantee the validity of the attribution statements.

One class of distributions relevant for attribution is the class of shift distributions $\mathcal{Q}$ generated by shift interventions \citep{Meinshausen2018} that are due to changes in the exogenous variables, i.e., one of the external forcings. More precisely, shift interventions are obtained by setting one of the exogenous variables to a given value $v$, leading to the class of distributions  
\begin{equation*}
\label{eq:shiftDistrib}
\begin{split}
	\mathcal{Q} =\ & \{ Q_v \text{ is the shifted distribution of } (X,Y)  \\ 
                & \text{ when the exogenous variable is set to a fixed value } v\}.    
\end{split}
\end{equation*}
In our case, we are interested in interventions affecting only $X$, and we make the assumption that the interventions considered will not affect the target variable $Y$, \ie the forcing of interest for attribution is not influenced by other factors. The population distribution $P$ from \eqref{eq:populationLoss} corresponds to the unshifted distribution of $(X,Y)$, \ie the case with no interventions. The class of distributions $\mathcal{Q}$ therefore contains the initial population distribution $P$, allowing the optimization over this larger class to lead to a more robust attribution. If one considers the different climate simulations (control runs, historical runs, etc., as discussed in Sect.~\ref{sect:detection}) as interventions corresponding to different distributions, then detection implicitly finds a robust fingerprint $\hat{\beta}$ for the set of distributions induced by these climate simulations.

Another way to view the fingerprint extraction from \eqref{eq:distrRobust} is as an optimization across heterogeneous environments \citep{Buehlmann2020, PetersEtAl2017},
\begin{equation*}
\label{eq:distrRobust_env}
\hat{\beta} = \operatorname*{argmin}_{\beta} \max_{e \in \mathcal{E}} \mathbb{E}_{(X^e, Y^e)} [\ell ( Y^e ,  f_{\beta}(X^e) ) ],
\end{equation*}
where $(X^e, Y^e)$ are samples from an environment $e \in \mathcal{E}$, and $\mathcal{E}$ is the space of (observed and unobserved) environments. Here, ``observed'' refers to the data available for analysis, \eg model simulations, reanalysis data or climate observations. Observed environments correspond to conditions where data has been collected or simulated, each observed environment being viewed as a subpopulation (or subset) of the available training data, while unobserved environments correspond to auxiliary potential conditions. In the motivating example from Sect.~\ref{sect:motivExample}, several climate simulations have been run with different values of the solar forcing (observed environments), however simulations for other values of the solar forcing would also be possible but have not been run (unobserved environments). An analogy can be made with the different Shared Socioeconomic Pathways \citep[SSPs,][]{RiahiEtAl2017}; if one would like to be robust to different values of the anthropogenic forcing when studying attribution to other forcings, one could include the different SSPs in the training data, but also potentially other counterfactual scenarios such as, for example, single-forcing simulations from the DAMIP project \citep{GilletEtAl2019}. 

In the framework of shift interventions, each environment $e$ corresponds to one distribution $Q_v$ from the class of distributions $\mathcal{Q}$ associated to a given shift $v$, $(X^e, Y^e) \sim Q_v$. Heterogeneity refers to the fact that the different environments correspond to different conditions, for example different solar forcings as in the motivating example from Sect.~\ref{sect:motivExample}. Distributional robustness guarantees that the learning model identifies those relationships between the climate variable $X$ and the target variable $Y$ that are invariant across the different environments considered. The space $\mathcal{E}$ could be constructed explicitly by running climate simulations with all the values of the shift $v$ that are relevant. All environments would then be observed and the fingerprints obtained would have invariance properties, however this process would be extremely computationally expensive and even infeasible if the goal would be to consider all the shifts within a given range. In the following, we describe anchor regression \citep{RothenhaeuslerEtAl2021} in the context of attribution. Anchor regression is a statistical method that allows us to consider a relevant space $\mathcal{E}$ of environments through the framework of shift interventions while also guaranteeing robustness properties. 

\section{Robust attribution using anchor regression}
\label{sect:anchorRegr}

Attribution of changes identified in a climate response $X$ to a specific forcing $F$ depends not only on $F$ but on all the other forcings that have a direct effect on $X$. To make the attribution robust we need to find a fingerprint that captures the climate response to the forcing $F$ but is robust to changes in other forcings, \ie the fingerprint representative of forcing $F$ should not change much (be invariant) if any of the other climate drivers change. Robust or invariant fingerprints make attribution less sensitive to reconstruction errors in the estimates of the external forcings and allow for variations between the training and test data, \eg between the climate models used for training and testing, or between the historical period and future projections of the climate. A robust attribution requires a prediction model that is robust to changes in the climate variable $X$ that are due to exogenous variables. In the example from Fig.\ \ref{fig:causal}b, when predicting the anthropogenic forcing $Y=F_3$, the exogenous variables are the solar forcing $F_1$ and the volcanic forcing $F_2$. We call these variables \textit{anchors}, and denote them by $A$. The exogenous variables contain information about the heterogeneity in the system and can be leveraged to extract invariant fingerprints. In the following, we present the anchor regression method in the context of attribution using so-called \textit{linear} anchors and discuss its connection to distributional robustness in Sect.~\ref{subsect:linearAnchor}. In Sect.~\ref{subsect:nonlinearAnchor}, we modify the anchor regression formulation to incorporate \textit{nonlinear} anchors and show the connection to mean independence, a lighter form of independence.

\subsection{Distributional robustness using anchor regression}
\label{subsect:linearAnchor}
Let us consider a linear setting with $f_{\beta}(X) = X \beta$. Given a set of $n$ pairs of observations $(\bm{x}_i, y_i)$, $i = 1 \ldots n$, and $f_{\beta}(\bm{x}_i) = \bm{x}_i \beta$, let the loss function $\ell$ used in \eqref{eq:populationLoss} be the least squares empirical risk,
\begin{equation}
\label{eq:leastSquares}
\mathbb{E}_{(X,Y)} [\ell ( Y ,  f_{\beta}(X) ) ] = \frac{1}{n}\sum_{i=1}^n \ell ( y_i ,  f_{\beta}(\bm{x}_i) ) = \frac{1}{n}\Vert \Y - \X \beta \Vert_2^2,
\end{equation}
where $\bm{x}_i \in \mathbb{R}^p$, $\bm{x}_i = (x_i^1, \ldots, x_i^p)$ is the $p$-dimensional input, $y_i \in \mathbb{R}$ is the target, and $\X \in \mathbb{R}^{n \times p}$ and $\Y \in \mathbb{R}^n$ are the input matrix and the output vector containing the $n$ observations used for training. The minimization of the least squares empirical risk ($L_2$-risk) from \eqref{eq:leastSquares} only considers the information contained in the variables $X$ and $Y$, despite the fact that $X$ is also influenced by $A$. If estimates of the anchor variables $A$ are available, one can additionally use the information contained in $A$ during training. We denote the anchor measurements by $\bm{a}_i \in \mathbb{R}^q$, where $q$ is the number of anchors. In the diagram from Fig.~\ref{fig:causal}b, $q=2$ for the two anchor variables, \ie the solar $F_1$ and volcanic $F_2$ forcings. We assume that we have estimates of the anchor variables for all $n$ pairs of observations, concatenated into the matrix $ \A \in \mathbb{R}^{n \times q}$. The training pairs of observations can then be extended to $(\bm{x}_i, y_i, \bm{a}_i)$. 

In the previous section we discussed one class of distributions relevant for attribution, that is, the class of shift distributions generated by shift interventions on $X$ that occur due to changes in the anchors. A recent statistical learning method called anchor regression \citep{RothenhaeuslerEtAl2021} has shown that one can achieve distributional robustness under shift interventions by adding a penalty term to the $L_2$-risk from \eqref{eq:leastSquares} that encourages invariance properties in the estimator $\hat{\beta}$, \ie the fingerprint, under changes in $A$. The robust estimator of anchor regression is given by
\begin{equation}
\label{eq:anchor}
\hat{\beta}^{\gamma} = \operatorname*{argmin}_{\beta} \Vert (I_n - \Pi_{\A})(\Y - \X \beta) \Vert_2^2 + 
					\gamma \Vert \Pi_{\A}(\Y - \X \beta) \Vert_2^2,		
\end{equation}
where $\Pi_{\A} \in \mathbb{R}^{n \times n}$ is the matrix that projects onto the column space of $\A$, \ie $\Pi_{\A} = \A(\A^T \A)^{-1} \A^T$, $I_n \in \mathbb{R}^{n \times n}$ is the identity matrix, and $\gamma$ is an intervention (or perturbation) strength hyperparameter. The second term is a penalty term which encourages the projection of the residuals from the prediction ($\Y -\X \beta$) onto the linear space spanned by the anchor to be as small as possible \citep[``near orthogonality'',][]{RothenhaeuslerEtAl2021}. Adding the penalty term corresponds to a least squares loss that minimizes the risk under shift interventions up to a given perturbation strength $\gamma$ \citep{RothenhaeuslerEtAl2021}. The new estimator $\hat{\beta}^{\gamma}$ is a distributionally-robust estimator as defined in \eqref{eq:distrRobust} for the class of shift distributions. In the example from Fig.~\ref{fig:causal}b, the solution of the regularized least squares from \eqref{eq:anchor} ensures good predictions for the anthropogenic forcing even if the solar or volcanic forcings change. Distributional robustness using anchor regression therefore allows for a robust attribution, that we call ``$\gamma$-attribution'', with the parameter $\gamma$ controlling the degree of robustness. 

The only constraint required for the anchor variables is that they are source (root) nodes in the causal graph in Fig.~\ref{fig:causal}, \ie there are no incoming edges in the graph for the anchors. In other words, the solar forcing $F_1$ and volcanic forcing $F_2$ are not directly influenced by other factors, a valid assumption for external radiative forcings. The variables $X, Y$, and $A$ are assumed to have zero mean, and therefore the matrices $\X, \Y$, and $\mathbf{A}$ are centered in a pre-processing step. There are three values of the parameter $\gamma$ that connect anchor regression to other known regression techniques \citep[see][ for more details]{RothenhaeuslerEtAl2021}: 1) for $\gamma = 0$, the estimator in \eqref{eq:anchor} coincides with partialling out the anchor variables and is optimized for prediction under zero perturbations, 2) for $\gamma = 1$, the estimator coincides with the ordinary least squares (OLS) estimator and protects only against very small perturbations $v$, and 3) for $\gamma \to \infty$, the estimator coincides with the two-stage least squares estimator used in instrumental variable (IV) regression \citep{BowdenTurkington1990}:
\begin{equation*}
\label{eq:gamma}
\begin{split}
\hat{\beta}^{0} & = \operatorname*{argmin}_{\beta} \Vert (I_n - \Pi_{\A})(\Y - \X \beta) \Vert_2^2,\\
\hat{\beta}^{1} & = \operatorname*{argmin}_{\beta} \Vert \Y - \X \beta \Vert_2^2,\\
\hat{\beta}^{\to \infty} & = \operatorname*{argmin}_{\beta} \Vert \Pi_{\A}(\Y - \X \beta) \Vert_2^2.
\end{split}
\end{equation*}

Anchor regression therefore interpolates between the solution to ordinary least squares ($\gamma = 1$) and two-stage least squares in instrumental variable regression ($\gamma \to \infty$). While OLS protects only against small shift interventions, IV regression protects against arbitrary strong interventions and identifies the true causal parameters that generate $X$ from $Y$. In an attribution setting, arbitrary strong perturbations are not justified as even in a changing climate, the increase in the magnitude of the external forcings and of their effect on $X$ remain bounded, and the true causal parameters are not necessarily needed for robust attribution. This is due to the fact that approximate invariance can already be achieved for values of $\gamma$ that are large enough (but not arbitrarily large) while also reducing the risk of trading-off too much predictive performance for robustness (more details on the choice of $\gamma$ in Sect.~\ref{subsect:nonlinearAnchor} and Sect.~\ref{sect:experiments}).

The anchor regression estimator from \eqref{eq:anchor} is prone to overfitting in high-dimensional settings where the problem is often ill-posed, \eg if the dimensionality of the data is larger than the sample size ($p > n$) as in the experimental settings presented in Sect.~\ref{sect:experiments} where $p$ is larger or of roughly the same magnitude as $n$. We therefore add a second regularization term to the optimization problem from \eqref{eq:anchor} to encourage smoothness of the fingerprints $\hat{\beta}$ for good generalization to unseen test data. We use ridge (Tikhonov) regularization \citep{HastieEtAl2009}, and rewrite the optimization problem from \eqref{eq:anchor} as ridge regression on a transformed dataset $(\tildeX, \tildeY)$ \citep[for more details see][]{RothenhaeuslerEtAl2021},
\begin{equation}
\label{eq:ridge}
\hat{\beta}^{\gamma} = \operatorname*{argmin}_{\beta} \Vert \tildeY - \tildeX \beta \Vert_2^2 + \lambda \Vert \beta \Vert_2^2,
\end{equation}
where $\lambda$ is the ridge regularization parameter, and $\tildeX = (I_n-\Pi_{\A}) \X+\sqrt{\gamma} \Pi_{\A} \X$ and $\tildeY = (I_n-\Pi_{\A}) \Y+\sqrt{\gamma} \Pi_{A} \Y$. The transformed dataset $(\tildeX, \tildeY)$ can be interpreted as artificially perturbed data for a given perturbation strength $\gamma$. Because perturbations are implicitly included in the training (without the need to explicitly generate them), this optimization allows to solve for a more flexible prediction problem than in \eqref{eq:populationLoss} that brings robustness to the attribution. The second term in \eqref{eq:ridge} penalizes large regression coefficients, encourages smoothness, and handles the multicollinearity of the predictors. The solution of the least-squares optimization in \eqref{eq:ridge} is the estimator 
\begin{equation}
\hat{\beta}^{\gamma} = (\tildeX^T \tildeX + \lambda I_n)^{-1} \tildeX^T \tildeY,
\label{eq:anchorClosedForm}
\end{equation}
which is biased but has reduced variance, thus avoiding overfitting to the training data. When $\lambda = 0$, $\hat{\beta}^{\gamma} = (\tildeX^T \tildeX)^{-1} \tildeX^T \tildeY$ is the unbiased OLS estimator on the transformed dataset. The choice of $\lambda$ controls the bias-variance trade-off, but also the trade-off between prediction accuracy and interpretability of the maps. The ridge $L_2$-penalty in \eqref{eq:ridge} encourages smoothness in the fingerprint, but depending on the application, the regularization term could be replaced by an $L_1$-penalty (LASSO) to encourage sparsity in the maps, an elastic net which is a linear combination of the $L_1$ and $L_2$ penalties to encourage both smoothness and sparsity, or penalties that take advantage of structural prior information about the climate system, \eg physical constraints \citep{BezenacEtAl2018, BeuclerEtAl2021} or graph-based regularizations \citep{BelkinEtAl2006} to account for geographical information.

Robustness on out-of-distribution data is usually achieved by trading off performance on unperturbed training data, \ie larger variance of the residuals at training time, for performance on potentially perturbed test data, \ie similar residual distribution at both training and testing times. Our experiments show however that the intervention regularization used in anchor regression can even improve the performance of the model on out-of-sample data compared to ridge regression while also providing robustness properties (see discussion in Sect.~\ref{sect:experiments}, Figs.~\ref{fig:GHG_aer_lin}-\ref{fig:aer_co2_nonlin}E).

\subsection{Connection between anchor regression and mean independence}
\label{subsect:nonlinearAnchor}

\begin{figure}
\center
	\includegraphics[width = \textwidth]{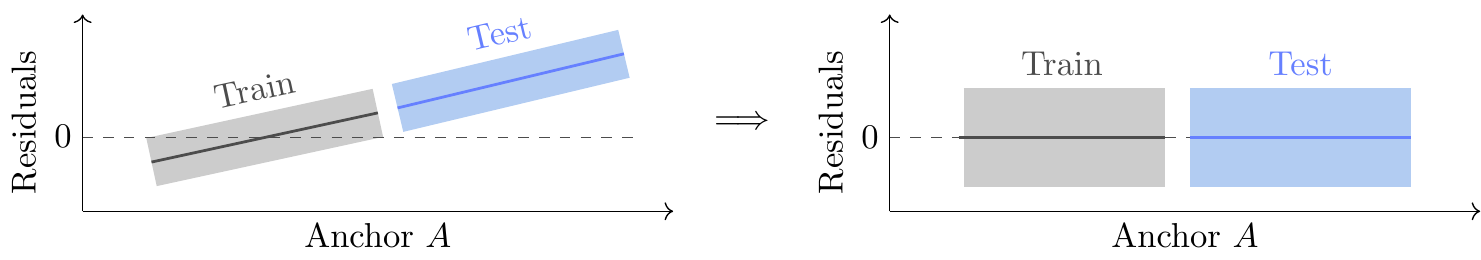}
	\caption{(Left) The residuals from the regression display a linear dependence on the anchor for the training data. If the anchor increases at testing time beyond the values observed during training, the residuals will also increase, leading to poor predictive performance. (Right) Using anchor regression, the linear dependence between the residuals and the anchor is removed in this example, which occurs for large values of the intervention strength hyperparameter $\gamma$. Larger values of the anchor variable at testing time compared to training do not affect the magnitude of the residuals. This is achieved by trading off predictive performance on unperturbed training data for predictive performance on perturbed data.}
\label{fig:residuals}
\end{figure}

Let the residual variable from the prediction be $R = Y - X\beta$. Through the penalty term in \eqref{eq:anchor}, anchor regression encourages the projection of the residuals on the linear span of the anchor variables $A$ to be as small as possible. This penalty is motivated by the fact that the anchors have a direct influence on both the climate variable $X$ and the residual variable $R$. An example of a linear dependence of the residuals on the anchor is depicted schematically in Fig.~\ref{fig:residuals}. Figure \ref{fig:residuals} (Left) shows how the residuals on the training data can have a small magnitude but be linearly dependent on the anchor. If at testing time the value of the anchor were to shift towards larger values not observed during training, the magnitude of the residuals would increase linearly with the anchor leading to a sub-optimal predictive performance. In the example from Fig.~\ref{fig:causal}b, if the solar or volcanic forcings were to increase, the prediction of the anthropogenic forcing would not be optimal and the attribution results might not remain valid. The idea behind anchor regression is illustrated in Fig.~\ref{fig:residuals} (Right) where the linear dependence between the residuals and the anchor has been removed. Even if the anchor were to increase significantly at testing time compared to training, the distribution of the residuals would remain the same. Anchor regression aims to find the direction $\beta$ that explains the component of the response of $X$ to the forcing of interest $Y$, \eg anthropogenic forcing, that is not shared in the climate response to other external forcings. In other words, the goal is to separate the effect of exogenous variables from the effect of the target on the climate variables. One way to achieve this separation is by reducing the part of the residuals that is explained by the exogenous variables, \ie the anchors. 

As defined in Sect.~\ref{subsect:linearAnchor}, anchor regression achieves distributional robustness by encouraging the projection of the residuals onto the linear space spanned by the anchor, $\Pi_{\A} = \A(\A^T \A)^{-1} \A^T$, to be as small as possible. We further call this case anchor regression with \textit{linear anchors}. The $L_2$-projection on the linear span of $A$ removes \textit{only} the linear dependence, \ie the correlation, between the residuals and the anchor, but does not remove other types of dependencies, \ie nonlinear dependencies such as periodicities. To remove nonlinear dependencies, we can modify the penalty term in anchor regression to project the residuals onto the space spanned by all the functions of $A$, not just linear functions. We further call this case anchor regression with \textit{nonlinear anchors}. From a geometrical perspective, the orthogonal projection ($L_2$-projection) of a random variable (here $R$) on the subspace of every possible function of $A$, \ie onto the whole space of $A$, is the conditional expectation $\mathbb{E}[R \mid A]$. We denote this projection by $P_A(R) = \mathbb{E}[R \mid A]$.

In a finite sample case, to obtain the projection $P_A$ we need to define a new column space $\Pi_{\A}$ containing both linear and nonlinear terms. Let the linear term be the identity, $h_1(A) = A$, and additional nonlinear terms be for example $h_2(A) = A^2, h_3(A) = |A|$, etc. We extend the column space $\Pi_{\A}$ used in \eqref{eq:anchor} to be the space spanned by all these functions. The anchor matrix becomes
\begin{equation*}
    \A_h = [h_1(\A)\ h_2(\A)\ h_3(\A)],
\end{equation*}
with corresponding column space $\Pi_{\A} = \A_h(\A_h^T \A_h)^{-1} \A_h^T$. Further adding more functions to $\A_h$ brings the column space closer to the true subspace covered by $A$ and the projection of the residuals onto this space closer to the conditional expectation.

The population version of anchor regression \citep{RothenhaeuslerEtAl2021} is defined as 
\begin{equation}
\label{eq:anchorPA}
\beta^{\gamma} = \operatorname*{argmin}_{\beta} \mathbb{E}_{\text{train}}[((I - P_A)(R))^2] + 
					\gamma \mathbb{E}_{\text{train}}[(P_A(R))^2],\\
\end{equation}
where $\mathbb{E}_{\text{train}}$ denotes the expectation over the training distribution. When the residuals are projected onto the whole space of $A$ not just the linear span, the $L_2$-projection $P_A$ becomes the conditional expectation as discussed above, $P_A(R) = \mathbb{E}[R \mid A]$, and the population version of anchor regression can be rewritten as (for the derivation see \ref{sect:ap_nonlinearAnchors})
\begin{equation}
\label{eq:anchorPA_Var}
\beta^{\gamma} = \operatorname*{argmin}_{\beta} \mathbb{E}_{\text{train}}[(R - \mathbb{E}_{\text{train}}[R \mid A])^2] + 
					\gamma \text{Var}(\mathbb{E}_{\text{train}}[R \mid A]).    
\end{equation}
Minimizing the projection of the residuals onto the space spanned by $A$ is thus equivalent to minimizing the variance of the conditional expectations $\mathbb{E}[R \mid A]$ on the training data, $\text{Var}(\mathbb{E}_{\text{train}}[R \mid A])$. In the limit of large data and $\gamma \to \infty$, the variance becomes equal to zero when all the conditional expectations become equal, \ie $\mathbb{E}_{\text{train}}[R \mid A = a]$ is constant for different values $a \in A$. This form of independence between two variables (here, $R$ and $A$) is called \textit{mean independence}. It requires that the first moment, \ie the mean or conditional expectation, does not depend on the other variable. In other words, $\mathbb{E}_{\text{train}}[R \mid A = a] = \mathbb{E}_{\text{train}}[R]$, $\forall a \in A$, \ie the expected value of the residuals for any value of the anchor should be the same. In addition, because the expectation of the residuals is always zero for the training data, $\mathbb{E}_{\text{train}}[R] = 0$, the constant conditional expectations would also converge towards zero, $\mathbb{E}_{\text{train}}[R \mid A = a] = 0$, leading to vanishing conditional expectations when $\gamma \to \infty$.

When the conditional expectation $\mathbb{E}[R \mid A]$ is a linear function of $A$, \eg such as multivariate Gaussian distributions for $(X, Y, A)$ and a linear setting \citep[for more details see][]{RothenhaeuslerEtAl2021}, the $L_2$-projection of the residuals on the linear span of $A$ is the conditional expectation $\mathbb{E}[R \mid A]$. Nonetheless, in more general cases and as discussed above, the $L_2$-projection on the linear span of $A$ removes only the linear dependence, \ie the correlation, between the residuals and the anchor, but does not lead to vanishing conditional expectations. To achieve mean independence, one would need to use anchor regression with nonlinear anchors as described above. The converse is however true: mean independence implies decorrelation between the residuals $R$ and the anchor $A$. Thus, while anchor regression with linear anchors achieves decorrelation and therefore robustness of the residuals to linear changes in the anchor, using nonlinear anchors further increases the robustness of the residuals to nonlinear changes.

In \ref{sect:ap_nonlinearAnchors}, we present more details on the motivation of using nonlinear functions to achieve vanishing conditional expectations and show how anchor regression achieves mean independence between the residuals and the anchor by minimizing the correlation ratio \citep{Renyi1959}, a measure of mean independence. In the experimental section, we show an attribution example where using a nonlinear anchor helps to bring the conditional expectations closer to zero in the prediction of the aerosols (Fig.~\ref{fig:aer_co2_nonlin}D) compared to using only the linear anchor (Fig.~\ref{fig:aer_co2_lin}D).

\section{Experiments}
\label{sect:experiments}

In this section, we first describe the climate data and the preprocessing steps, followed by the training and testing procedure, and finally, we present the results for two attribution cases.

\subsection{Data processing}

Let $\X \in \mathbb{R}^{N \times p}$ denote the matrix of climate data, and $\Y \in \mathbb{R}^N$ the target vector, both from model simulations. Here, $\X$ is obtained by concatenating the different model simulation runs, and $N$ denotes the total number of samples, \ie the total number of simulated years across all simulations. Concatenation allows us to also take into account aspects related to internal climate variability and model structural differences across climate models in addition to the mean pattern. The total number of samples is $N=s \times u$, where $s$ is the number of model simulation runs and $u$ the number of time steps per model run, \eg here years, assuming the same number of simulated years for each run. We choose the target vector $\Y$ to reflect an estimate of global-scale effective radiative forcing, obtained from a simple energy balance model \citep{SmithEtAl2021}, which includes estimates of both GHG-induced and aerosols-induced radiative forcing. In more traditional so-called ``non-optimal'' D\&A, a metric that reflects the forced pattern, \ie the fingerprint, and a forced signal are determined by a step similar to averaging multiple model simulations to reduce internal variability and subsequent EOF analysis \citep{SanterEtAl1996, SanterEtAl2019}.

We train the statistical model using both forced (historical) and unforced (control runs) model simulations from the CMIP6 (Climate Model Intercomparison Project) multi-model archive \citep{EyringEtAl2016}. The historical runs span the time period from 1850--2014 ($u = 165$ simulated years per run), and for control runs we use the same number of years per run, \ie $u=165$, to give a balanced importance to forced and unforced runs in the training of the statistical model. For the control runs we chose the last $165$ years of the unforced simulations to allow convergence of the climate models and to remove the effect of the initial conditions. The full dataset consists of $s = 102$ runs (52 historical runs and 50 control runs) from a set $\mathcal{M}$ of $|\mathcal{M}| = 27$ models from different modelling centers (all models used and their variants are shown in \ref{sect:ap_models}) with a total of $N = s \times u = 102 \times 165 = 16,830$ samples, \ie total number of simulated years. The samples are two-dimensional spatial maps regridded to a common regular $2.5^{\circ} \times 2.5^{\circ}$ spatial grid for all simulations, leading to a spatial dimension $p = 144 \times 72 = 10,368$ grid cells. The mean of each grid cell of the first 50 years is subtracted from each model run individually in order to remove model biases in mean temperature, and the data is standardized prior to the regression analysis. 

\subsection{Training and testing procedure}
\label{sect:train_test_proc}

We train anchor regression using half of the models (with a total of $n$ number of samples, see Sect.~\ref{sect:anchorRegr}) to obtain the fingerprint, while the other half of the models ($N-n$ number of samples) are used as held-out test data for prediction and hypothesis testing. For every climate model we use all the runs available, \ie in our case, historical and control runs. We repeat the procedure $B=50$ times for different model train-test splits to obtain multiple fingerprints $\hat{\beta}_b$, $b=1 \ldots B$, that are then aggregated using subagging \citep[subsample aggregating,][]{BuehlmannYu2002} to reduce the variance of the final estimator for the fingerprint $\hat{\beta} = \frac{1}{B}\sum_{b=1}^B \hat{\beta}_b$. Two hyperparameters need to be tuned in anchor regression, i.e, the intervention strength hyperparameter $\gamma$ and the ridge regularization hyperparameter $\lambda$. Within each of the $B$ subsamples we use $K$-fold cross validation ($K=3$ folds) to select the optimal hyperparameters using a grid search over 9 candidate $\gamma$ values ($\gamma \in \{1, 2, 5, 10, 10^2, 10^3, 10^4, 10^5, 10^6 \}$) and 50 candidate $\lambda$ values selected on a logarithmic scale ($\lambda \in [1, 10^9]$). Each of the $K$ folds is left out sequentially as the validation set, and the fingerprints $\hat{\beta}_b^{\gamma, \lambda}$ for each ($\gamma$, $\lambda$) pair are obtained by averaging the $K$ fingerprints from the cross validation, \ie $\hat{\beta}_b^{\gamma, \lambda} = \frac{1}{K}\sum_{k=1}^K \hat{\beta}_{b,k}^{\gamma, \lambda}$. The fingerprint $\hat{\beta}_b$ corresponding to the optimal values of the hyperparameters (selected as described below) will be used for prediction and hypothesis testing on the held-out test data. If modelling centers provide multiple variants of a particular model, these variants are all included either in the training or test datasets, and similarly in the same fold in cross validation, in order to avoid any biases due to model overlaps. 

The selection of the hyperparameters depends on two objectives: 1) minimization of the total prediction error, \ie the root mean squared error ($RMSE$), and 2) minimization of the component of the $RMSE$ that correlates with the anchor, for robustness properties. The latter will be denoted as $RMSE_{\Pi_{\A}}$ and accounts for the part of the residuals that projects onto the space spanned by the anchor $\Pi_{\A}$. While the goal is to reduce both $RMSE$ and $RMSE_{\Pi_{\A}}$ simultaneously, a trade-off appears between the two objectives as an improvement of one often leads to a deterioration of the other. The optimal hyperparameters ($\gamma$, $\lambda$) within each model train-test split subsample are therefore selected using a multiobjective optimization criterion \citep{Miettinen1998} which captures the trade-off as a Pareto optimal front where none of the objectives can be improved without deteriorating the other (lower left corner of Figs.~\ref{fig:GHG_aer_lin}--\ref{fig:aer_co2_nonlin}E). We use a weighted sum of the normalized objectives above for the multiobjective optimization, and more specifically, the weighted $L_2$-problem \citep[][p.~97]{Miettinen1998},
\begin{equation}
    \min \left[ \sum_{j=1}^C w_j \left( \frac{\phi_j(\gamma, \lambda) - z_j^*}{z_j^{\text{nad}} - z_j^*} \right)^2 \right]^{\frac{1}{2}},
\end{equation}
where $C$ is the number of objectives, $\phi_j(\cdot)$ are the objectives as functions of $\gamma$ and $\lambda$, $w_j$ are the importance weights assigned to each objectives, and $z_j^*$ and $z_j^{\text{nad}}$ are the ideal and nadir objective vectors. The ideal objective vectors $z_j^*$ would achieve minimization of each objective individually but are unachievable in practice, while the nadir objective vectors $z_j^{\text{nad}}$ capture the worst individual objective values along the Pareto front. The weights are chosen to be positive and to sum to one, \ie $w_j \geq 0$ and $\sum_{j=1}^C w_j = 1$. 

Once the optimal hyperparameters are selected for each train-test split subsample, we use the corresponding optimal fingerprint $\hat{\beta}_b$ for that subsample to make predictions of the target $Y$ on the held-out models and test attribution. Each climate model $m$ from the set of held-out models within the subsample is left out iteratively, and each of its runs (control and/or historical) is then compared against the null distribution of the control runs of the held-out models excluding model $m$ as described in Sect.~\ref{sect:test_statistics} and \ref{sect:HT_appendix}. Attribution to the target forcing $Y$ occurs if the test statistic, \ie the rank correlation between the values of the predicted forcing and the actual forcing (see Sect.~\ref{sect:test_statistics}), falls in the critical region of the null distribution for a two-tailed test with a significance level $\alpha_* = 0.05$ (more details in \ref{sect:HT_appendix}). The procedure is repeated $B$ times and the results of the hypothesis testing are aggregated (see \ref{sect:HT_appendix}) to estimate the type I error and power for each model. For detection, these measures estimate how often unforced control runs are mistaken as externally-forced (type I error) and how often externally-forced runs such as historical runs are correctly identified as being externally-forced (power). For attribution, the type I error measures how often we wrongly attribute a change to a given forcing when the respective forcing is actually not present, and the power measures how often a given forcing has been correctly identified as being present based on the estimated fingerprint. If the observations are considered as one of the models, then the procedure is straightforward to apply to observations. If the statistical model is accurate, \ie produces a test with a low type I error (close to the chosen significance level $\alpha_*$) and a high power (close to one), then a test statistic for the observations that falls in the critical region of the null distribution indicates that the observations are externally-forced in the case of detection or forced with the forcing of interest in the case of attribution.

\subsection{Results}
\label{sect:results}

In the following, we investigate two attribution scenarios for temperature within the CMIP6 models: 1) attribution to the GHG forcing while protecting against changes in the aerosols, and 2) attribution to the aerosols forcing while protecting against changes in the CO$_2$ forcing. The radiative forcings are shown in Fig.~\ref{fig:radForcings} in \ref{sect:ap_forcing_corr}.

\subsubsection{Attribution to the GHG forcing with the aerosols anchor}
\label{sect:GHG_aer}

\begin{figure}
\centering
    \includegraphics[width = 0.95\textwidth]{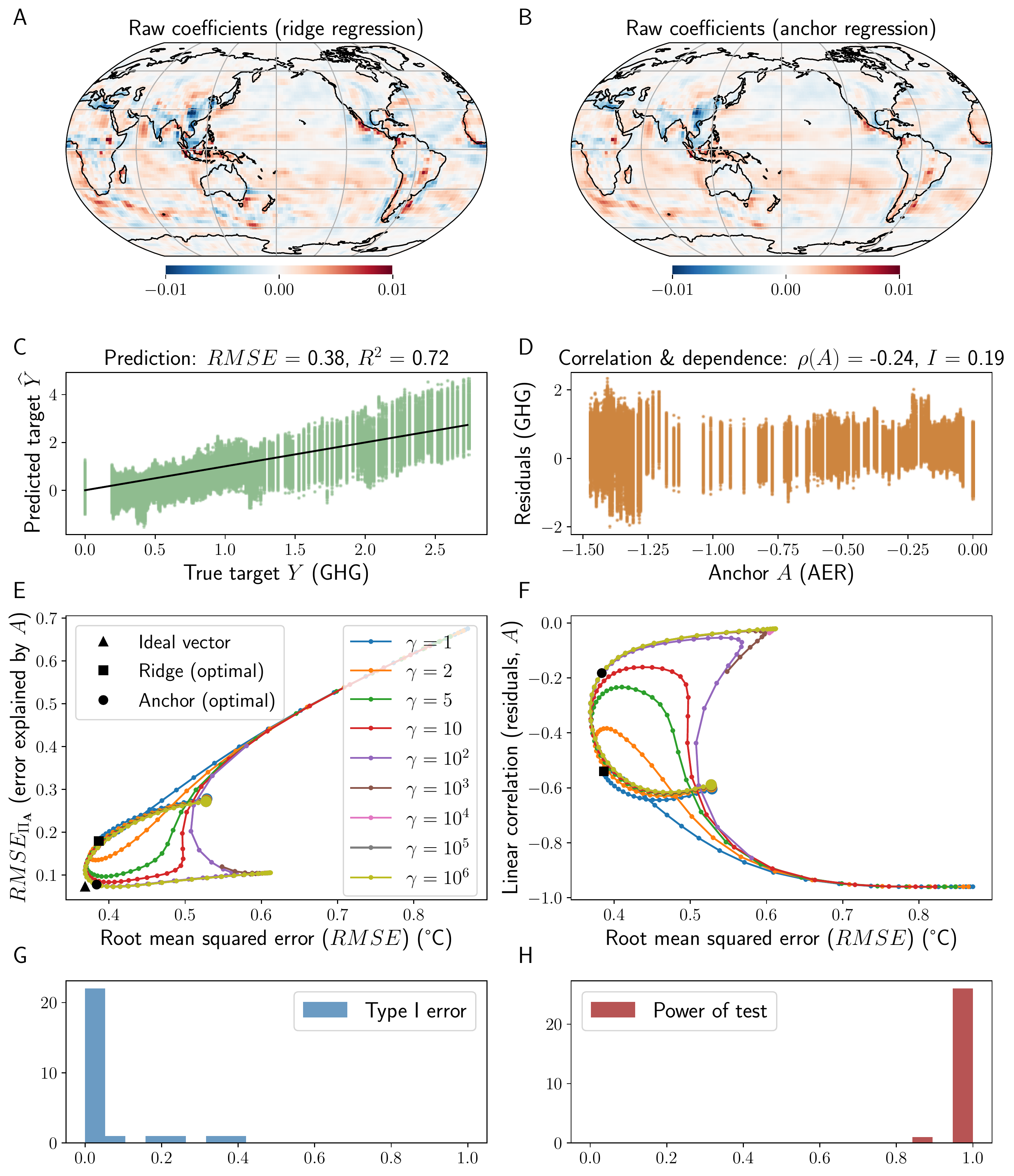}
	\caption{Attribution of the GHG forcing with the aerosols (linear) anchor. (A) Optimal ridge fingerprint (black square in Panel E). (B) Optimal anchor fingerprint (black circle in Panel E). (C) Predictions for the GHG forcing using the anchor fingerprint from Panel B. (D) Correlation and mutual independence of the residuals from the predictions with the aerosols anchor. (E) $RMSE$ vs.~$RMSE_{\Pi_{\A}}$ for different values of the intervention strength and ridge regularization hyperparameters ($\gamma$, $\lambda$). (F) $RMSE$ vs.~linear correlation of the residuals with the anchor for the same hyperparameters ($\gamma$, $\lambda$) as in (E). (G, H) Type I error and power. 
	\label{fig:GHG_aer_lin}}
\end{figure}

We first predict and test attribution to the GHG forcing (forcing of interest $F$) using the aerosols as anchor ($A$). This is a highly relevant practical problem in climate science because knowledge of the forced GHG response in historical simulations (and by extension, observations) would potentially allow to derive more targeted observational constraints on future GHG-induced warming or transient climate response \citep[for more details see, \eg][]{NijsseEtAl2020,TokarskaEtAl2020}.
The GHG forcing is computed as the sum of the CO$_2$, CH$_4$ and N$_2$O forcings, and the results using anchor regression are presented in Fig.~\ref{fig:GHG_aer_lin}. Panel A shows the estimated optimal ridge fingerprint $\hat{\beta}$ as the raw coefficients from ridge regression averaged over the $B=50$ train-test splits (see Sect.~\ref{sect:train_test_proc}). A different value of the hyperparameter $\lambda$ is selected for each train-test split by the optimization procedure. Ridge regression is equivalent to anchor regression with hyperparameter $\gamma=1$ therefore there is no optimization over $\gamma$. Panel B shows the estimated optimal anchor fingerprint as the averaged raw coefficients from anchor regression for the optimal hyperparameters $\gamma$ and $\lambda$ over the $B$ train-test splits. The most common value selected for the intervention strength hyperparameter is $\gamma=10^6$ (for 47 out of the 50 train-test splits). The two optimal fingerprints (ridge and anchor) are obtained using the weighted multiobjective optimization criterion from Sect.~\ref{sect:train_test_proc} with $C=2$ and weights $w_1=0.5$ for $RMSE$ and $w_2=0.5$ for $RMSE_{\Pi_{\A}}$, thus giving equal weight to the two normalized objectives. While the anchor fingerprint is slightly smoother than the ridge fingerprint, both fingerprints assign a positive contribution (positive raw coefficients) to the tropics (except for the El-Ni\~no region) and Southern Hemisphere's oceans in attributing changes in temperature to the GHG forcing while protecting against changes due to the aerosols. Other regions might also positively contribute to explaining the GHG forcing, but if they also contribute to explaining the aerosols forcing, then anchor regression does not highlight those regions as it tries to increase the robustness of the predictions to changes due to the aerosols forcing. Negative coefficients are assigned mostly to the Northern Hemisphere with the strongest emphasis on South-East Asia, India and the Mediterranean Sea, regions that are weakly correlated with the GHG forcing in the historical record and for some, such as South-East Asia and inland United States, also positively correlated with the aerosols anchor (the correlation maps of temperature with the GHG, CO$_2$ and aerosols forcings are shown in Fig.~\ref{fig:corr_forcings} in \ref{sect:ap_forcing_corr}). Negative coefficients reduce the degree to which the temperature response to the aerosols forcing projects onto the anchor fingerprint. 

Panel C shows the prediction results for anchor regression with an $RMSE =0.38$ and a coefficient of determination $R^2=0.72$. Panel D shows the correlation and dependence (mutual information) of the residuals from anchor regression with the anchor variable. Both the correlation, $\rho(A)=-0.24$, and dependence (measured by mutual information), $I = 0.19$, are low when using the optimal fingerprints from anchor regression. Panels C and D show results on held-out test data, but knowledge about the anchor is only needed at training time. The choice of the optimal hyperparameters ($\gamma$, $\lambda$) is shown in Panel E on the validation data. The different curved lines correspond to different values of the intervention strength hyperparameter $\gamma$. Each line starts at the large corresponding circle for the ridge regularization hyperparameter $\lambda = 1$ which then increases on a logarithmic scale in the range $\lambda \in [1, 10^9]$. The blue line ($\gamma = 1$) corresponds to ridge regression, and as $\gamma$ increases for anchor regression, $RMSE_{\Pi_{\A}}$ decreases significantly (up to a reduction of 50$\%$ for large $\gamma$ values). One important aspect observed in Panel E is that anchor regression achieves a lower $RMSE$ compared to ridge regression for all $\gamma$ values on the validation data due to the robustness introduced by the penalty term. Anchor regression therefore provides an additional regularization on the residuals without deteriorating the predictive performance on the data. The optimal hyperparameters for ridge and anchor regression are chosen as the points on the lines closest to the ideal vector (black triangle) using the normalized objectives. The optimal anchor solution (black circle, fingerprint in Panel B) has roughly the same $RMSE$ as the optimal ridge solution (black square, fingerprint in Panel A) but a significantly lower projection of the residuals onto the space spanned by A as captured by $RMSE_{\Pi_{\A}}$ ($RMSE_{\Pi_{\A}} \approx 0.2$ for ridge regression vs.~$RMSE_{\Pi_{\A}} \approx 0.1$ for anchor regression). Anchor regression therefore achieves a significant gain in robustness to changes due to the anchor for a similar $RMSE$. Both $RMSE$ and $RMSE_{\Pi_{\A}}$ decrease as $\gamma$ and $\lambda$ increase but only up to a given point when only $RMSE_{\Pi_{\A}}$ continues to decrease significantly. As $\gamma$ increases, anchor regression gives more and more importance to $RMSE_{\Pi_{\A}}$ which remains low, while $RMSE$ continues to increase for large values of the ridge hyperparameter $\lambda$. The sudden increase in $RMSE_{\Pi_{\A}}$ at the end of the curves is due to the fact that a strongly ridge regularized model (large $\lambda$) brings the predicted target $\hat{Y}$ close to zero and therefore the residuals close to the target $Y$. If the projection of $Y$ (and therefore of the residuals) onto the space spanned by the anchor $A$ is large (here, GHG and aerosols are strongly negatively correlated) then $RMSE_{\Pi_{\A}}$ also increases. Panel F shows the Pearson linear correlation of the residuals with the anchor for the same ($\gamma, \lambda$) pair values as in Panel E. The optimal ridge and anchor solutions are the same ones as in Panel E. Even though correlation is not an objective that is explicitly minimized by anchor regression, the anchor solution achieves a significant reduction in correlation, \ie more than 50$\%$ with respect to ridge regression. Thus, $\rho(A) \approx -0.55$ for ridge regression (black square) vs.~$\rho(A) \approx -0.2$ for anchor regression (black circle) in Panel F. The results for the evaluation metrics on the held-out test data shown in Fig.~\ref{fig:GHG_aer_lin_test} in \ref{sect:ap_addResults} are very similar to the results obtained on the validation data. 

Panels G and H show the results of the hypothesis testing for a two-tailed test with a significance level $\alpha_* = 0.05$ (see details in \ref{sect:HT_appendix}), \ie a $95\%$ confidence level or a threshold that is 1.96 standard deviations away from the mean for a standard normal Gaussian. Most models have a low estimated type I error, as follows: 19 out of the 27 models have no type I error, $\hat{\alpha}_m = 0$; four models have $\hat{\alpha}_m < 0.1$; two models have $\hat{\alpha}_m \approx 0.2$ (Nor, SAM); and two models have $\hat{\alpha}_m \approx 0.4$ (CMC, GIS). Low type I errors indicate that the control runs from the corresponding models have not been mistakenly identified as externally-forced with the GHG forcing. Almost all models (23 out of 27) achieve the maximum power, $\hat{\kappa}_m=1$, with one model (CAS) having a power $\hat{\kappa}_m \approx 0.9$ and three models with $\hat{\kappa}_m > 0.95$. High power indicates that for runs externally-forced with the GHG forcing (here, historical runs), (part of) the changes were correctly attributed to the GHG forcing. The average type I error across models is  $\hat{\alpha} = 0.0496$, close to the significance level chosen $\alpha_* = 0.05$, and the average power across models is $\hat{\kappa} = 0.9934$, close to one. Low type I error and high power give confidence in the attribution of the changes to the GHG forcing while protecting against other forcings that can interfere with the target, such as the aerosols in this case. We also show results for the attribution to the GHG forcing using nonlinear anchors in Fig.~\ref{fig:GHG_aer_nonlin} in \ref{sect:ap_addResults}. In the prediction of the GHG forcing, using the nonlinear aerosols anchor slightly increases $RMSE$ from $RMSE =0.38$ to $RMSE = 0.39$ (more details on the interpretation of the nonlinear anchors in Sect.~\ref{sect:aer_co2}) while the dependence remains the same (mutual information $I=0.19$). The results of the hypothesis testing are similar to the linear anchor with a type I error $\hat{\alpha} = 0.0573$ and power $\hat{\kappa} = 0.9952$. We repeated the bagging procedure for the linear and nonlinear cases and the results are very robust with the same models having a high type I error or low power.

\subsubsection{Attribution to the aerosols forcing with the CO$_2$ anchor}
\label{sect:aer_co2}

\begin{figure}
    \includegraphics[width = 0.95\textwidth]{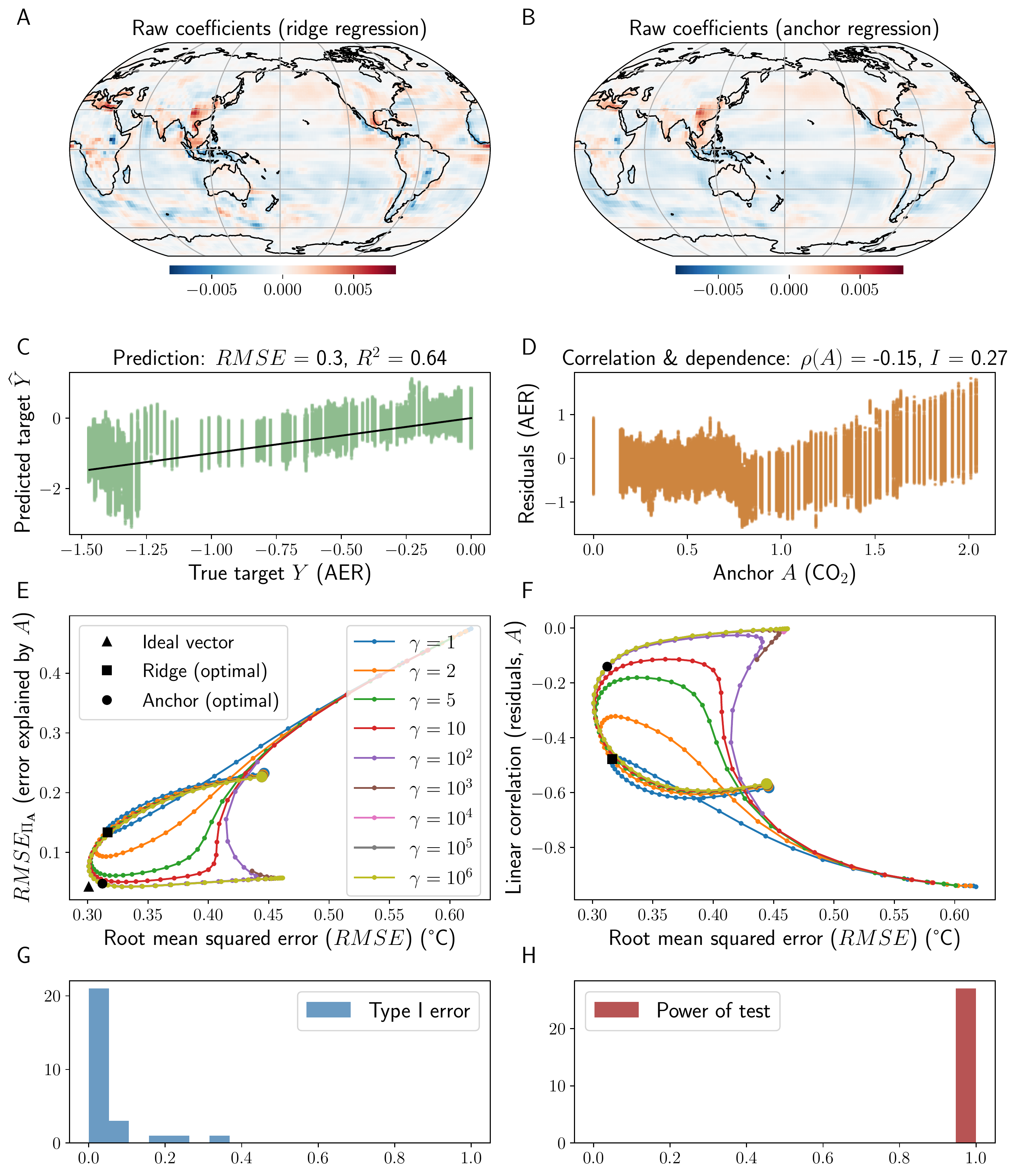}
	\caption{Prediction and attribution of the aerosols forcing with the CO$_2$ linear anchor. Same panels as in Fig.~\ref{fig:GHG_aer_lin}. 
	\label{fig:aer_co2_lin}}
\end{figure}

We now test attribution to the aerosols forcing using the CO$_2$ anchor, and display the results in Fig.~\ref{fig:aer_co2_lin} for the linear anchor. The ridge (Panel A) and anchor (Panel B) fingerprints show that the Northern Hemisphere is assigned mostly positive coefficients therefore a positive contribution in explaining the aerosols forcing while protecting against the CO$_2$ forcing, with a strong emphasis on South-East Asia and the Mediterranean Sea. The aerosol fingerprint is very similar in pattern to the GHG fingerprint but negatively correlated with it (see also the correlation maps of temperature with the GHG and aerosols forcings in Fig.~\ref{fig:corr_forcings}; for the GHG forcing, the correlation is always positive, and for the aerosols forcing, the correlation is mostly negative with the exception of South-East Asia and a small area of inland United States which have low positive correlations). Despite the fact that the aerosols pattern changes over time due to a change in emission locations, the statistical model is able to find an aerosols fingerprint that leads to good predictions while also protecting against changes in the CO$_2$ forcing. Thus, the prediction error is $RMSE = 0.3$ and $R^2 = 0.64$ (Panel C) with a correlation between the residuals and the anchor of $\rho(A)=-0.15$ and a mutual information of $I=0.27$ (Panel D). As opposed to the prediction of the GHG forcing, we observe a clear dependence (here, almost a linear trend) of the residuals on the anchor starting with a value of the anchor of $A \approx 0.75$. This dependence explains the large mutual information despite a low correlation. The ideal vector, and the optimal ridge and anchor solutions on the validation data are shown in Panel E. Similarly to the GHG forcing experiment, anchor regression leads to a significant reduction in $RMSE_{\Pi_{\A}}$ compared to ridge regression without an increase in $RMSE$. The most common value chosen for the intervention strength hyperparameter is $\gamma = 10^6$ (45 out of the 50 train-test splits), the largest among the values considered and the same as in the case of the attribution to the GHG forcing. However, despite the fact that the largest $\gamma$ value is chosen, we observe in Panel E that the $RMSE$ and $RMSE_{\Pi_{\A}}$ values corresponding to $\gamma \geq 10^2$ are not distinguishable in the lower left corner (the relevant values for optimization) and therefore any of these values of $\gamma$ would lead to similar results in terms of performance. However, higher values of $\gamma$ correspond to stronger potential interventions and can therefore guarantee robustness to larger future changes due to the anchor. The results on the test data are shown in Fig.~\ref{fig:aer_co2_lin_test} in \ref{sect:ap_addResults}, and are very similar to the results on the validation data. The results of the hypothesis testing are shown in Panels G and H. The type I error $\hat{\alpha}_m = 0$ for 19 out of the 27 models, with one model having an error of $\hat{\alpha}_m \approx 0.3$ (GIS), two models an error of $\hat{\alpha}_m \approx 0.2$ (CMC, Nor), and five models an error of $\hat{\alpha}_m < 0.1$. The average type I error across models is $\hat{\alpha} = 0.0376$ for a significance level of $\alpha_* = 0.05$. All models (27 out of 27) achieve the maximum power, \ie $\hat{\kappa}_m = 1$, and the average power across models is therefore $\hat{\kappa} = 1$. We observe that the models with high type I error for the attribution to the GHG forcing are roughly the same models with high type I error for the attribution to the aerosols forcing.

\begin{figure}
    \includegraphics[width = 0.95\textwidth]{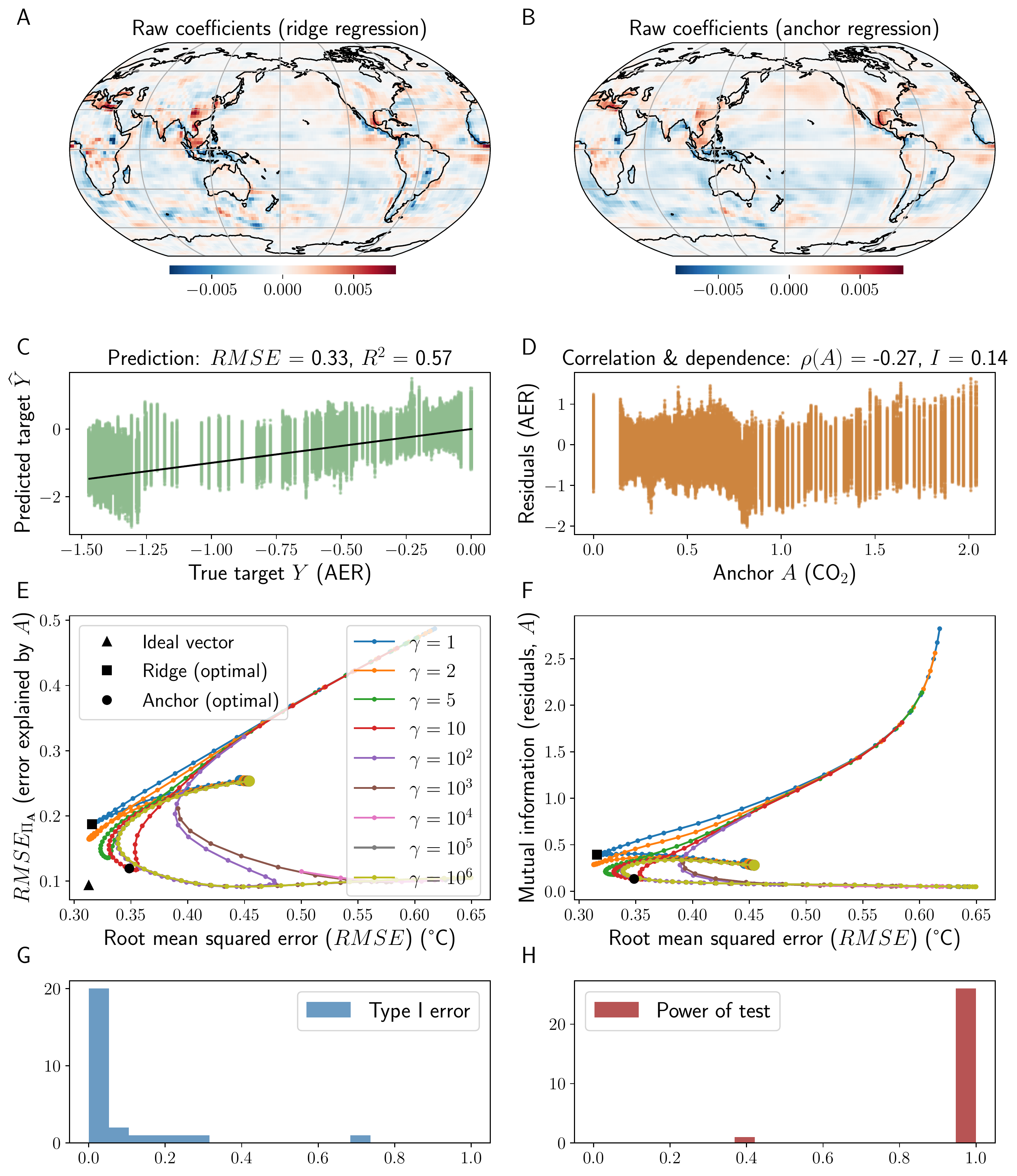}
	\caption{Prediction and attribution of the aerosols forcing with CO$_2$ nonlinear (square) anchors. Panels similar to Fig.~\ref{fig:GHG_aer_lin}. Nonlinear anchors help to further reduce the dependence (mutual information) of the residuals with the anchor compared to the linear anchor (see Fig.~\ref{fig:aer_co2_lin}D).
	\label{fig:aer_co2_nonlin}}
\end{figure}

Panel F in Figs.~\ref{fig:GHG_aer_lin} and \ref{fig:aer_co2_lin} indicates that anchor regression with the linear anchor is able to remove a much larger part of the linear dependence, \ie linear correlation, between the residuals and the anchor compared to ridge regression. Despite the fact that the correlation is not an objective that is minimized explicitly in anchor regression, there is an implicit connection between them as discussed in \ref{sect:ap_nonlinearAnchors}. However, other nonlinear dependencies remain that can bias the attribution, as shown in Fig.~\ref{fig:aer_co2_lin}D where the mutual information remains relatively large ($I=0.27$). To further reduce dependencies, we use anchor regression with nonlinear anchors. More specifically, we add the square anchor to the linear anchor and use $\A_{h} = [\A, \A^2]$ to compute the projection matrix $\Pi_{\A}$ (see Sect.~\ref{subsect:nonlinearAnchor}). The results are displayed in Fig.~\ref{fig:aer_co2_nonlin}. Panel D shows that anchor regression with the nonlinear anchor is able to further remove nonlinear dependencies leading to a halved mutual information $I=0.14$ (as opposed to $I=0.27$ for the linear anchor, Fig.~\ref{fig:aer_co2_lin}D) despite an increase in the linear correlation $\rho(A) = -0.27$ (as opposed to $\rho(A) = -0.15$ for the linear anchor, Fig.~\ref{fig:aer_co2_lin}D) and a slight increase in $RMSE = 0.33$ (as opposed to $RMSE = 0.3$ for the linear anchor, Fig.~\ref{fig:aer_co2_lin}C). Further adding nonlinear anchors, \ie nonlinear functions of $A$, to $\A_{h}$ will further decrease the dependence by expanding the space covered by functions of $A$ (see Sect.~\ref{subsect:nonlinearAnchor} and \ref{sect:ap_nonlinearAnchors} for futher discussion). The nonlinear anchor fingerprint is very similar to the linear anchor fingerprint, however the Northern Hemisphere is assigned slightly more balanced contributions without the strong emphasis on South-East Asia as for the linear anchor, and the Gulf of Mexico is assigned relatively large coefficients. The optimal hyperparameters ($\gamma, \lambda$) are chosen using the weighted multiobjective optimization with two objectives ($C=2$): $RMSE$ and $RMSE_{\Pi_{\A}}$ with weights $w_1=0.5$ and $w_2=0.5$, respectively. Panel E displays the $RMSE$ vs.~$RMSE_{\Pi_{\A}}$. The optimal anchor solution (black circle) is obtained for much lower values of $\gamma$ compared to the linear anchor (lower left corner of Panel E). This is due to the fact that adding nonlinear functions of the anchor expands the projection space and therefore the amount of intervention strength regularization required is smaller. The most common value chosen for the intervention hyperparameter is $\gamma = 10$ (36 out of the 50 train-test splits) as opposed to $\gamma=10^6$ for the linear anchor. For the nonlinear anchor, Panel F shows the mutual information (instead of the correlation) vs.~$RMSE$ where we see that the anchor solution achieves a much lower dependence (black circle, $I \approx 0.1$) compared to the ridge solution (black square, $I \approx 0.4$). The results on the test data are shown in Fig.~\ref{fig:aer_co2_nonlin_test} in \ref{sect:ap_addResults}. The hypothesis testing (Panels G and H) reveals a slightly higher type I error and a lower power for the nonlinear anchors compared to the linear anchor. Removing nonlinear dependencies with the anchor comes at a slight cost in accuracy, but guarantees a more robust attribution to the forcing of interest. The average type I error across models is $\hat{\alpha} = 0.0616$ for a significance level of $\alpha_* = 0.05$, and the average power $\hat{\kappa} = 0.9767$. Most of the models (20 out of 27) have no type I error, four models have an error of $\hat{\alpha}_m \approx 0.1$ (ACC, CES, GFD, MPI), two models $\hat{\alpha}_m \approx 0.2$ (GIS, Nor), and one model $\hat{\alpha}_m \approx 0.7$ (NES). Almost all models (25 out of 27) have the maximum power of one, with only one model having low power $\hat{\kappa}_m \approx 0.4$ (CMC).

\section{Conclusion}

We introduced in this paper a novel approach for direct detection and attribution of climate change that relies on supervised learning and distributional robustness, and therefore allows for a robust detection and attribution. Direct D\&A takes explicitly into account the information contained in the target of interest, and finds fingerprints that are the most representative of the respective targets. We employ a statistical learning method inspired by causal inference, namely anchor regression \citep{RothenhaeuslerEtAl2021}, to guarantee robustness to changes in forcings other than the target, so-called anchor variables. The fingerprints extracted through anchor regression capture features that are invariant to changes in the anchor variables across a set of model simulations, and therefore have robustness properties. Both detection and attribution use the same statistical learning model but with different target variables. While detection uses as target a metric that encapsulates the total external forcing, attribution uses as target the forcing of interest, \eg anthropogenic forcing, aerosols forcing, or a related metric. 

We present results for two attribution scenarios for the GHG and aerosols forcings while anchoring, \ie protecting, against changes due to the aerosols and CO$_2$ forcings, respectively. The prediction for the two targets shows high accuracy, \ie low $RMSE$, and robustness to changes in the anchor, \ie small projection of the residuals onto the space spanned by the anchor. Distributional robustness is achieved through the penalty term in anchor regression that acts as a regularization and reduces the effect of uncertainties in the source of the forcing or to future changes in the external forcings. A second ridge penalty term helps to give lower weight to regions with large internal variability or large disagreement between climate models, thus improving the signal-to-noise ratio, similar to signal-to-noise optimization in traditional D\&A \citep{Hasselmann1979, HegerlEtAl1996}. In some of the experiments, we obtain a lower $RMSE$ for anchor regression compared to ridge regression despite the fact that an additional penalty term is present, showing the good extrapolation capabilities of anchor regression. Using only the historical time period from 1850 to 2014 in the experiments, we are able to attribute part of the past changes in temperature to both GHG and aerosols forcings. Despite the fact that correlation and nonlinear dependence measures, such as mutual information, are not explicitly minimized in anchor regression, reducing the projection of the residuals onto the space spanned by the anchors also significantly reduces the correlation and nonlinear dependence compared to ridge regression. We show that this behaviour is explained by the implicit connection between the loss function of anchor regression and mean independence through the connection to the correlation ratio defined in \citet{Renyi1959} (\ref{sect:ap_nonlinearAnchors}).
 
Our D\&A methodology was shown to work well in a ``model as truth'' setting where the statistical model is tested on a set of climate models not seen during training, and will provide the basis for D\&A on observations in future work. In the motivating example in Sect.~\ref{sect:motivExample}, changes in the solar forcing do not affect the prediction of CO$_2$ if one is anchoring against the solar forcing within a single model large ensemble. The direct D\&A is framed in a high-dimensional supervised linear setting, but nonlinear supervised methods, such as neural networks, could be used as well as long as they incorporate the information on the exogenous variables during training. In the linear setting, the fingerprint is represented by the regression coefficients that predict, across a set of climate model simulations, a specific forcing or a metric that captures the forced response, \eg anthropogenic forcing, from a map of climate variables, \eg temperature, precipitation. The predicted target, which is often one-dimensional, is used to define a test statistic for assessing detection and attribution. As our approach is based on supervised learning there is no projection onto the subspace spanned by the first few EOFs as in traditional D\&A, thus avoiding the problem of imposing orthogonality of the representative fingerprints. 

The work presented here opens new avenues on using supervised statistical learning and machine learning to increase the robustness of D\&A to changes in external forcings other than the target of interest or on protecting against changes in internal variability \citep{SippelEtAl2021}. Multiple climate variables can easily be incorporated into anchor regression, \eg temperature and precipitation, with promising initial results that are however beyond the scope of the current paper.

\acks{We acknowledge the World Climate Research Programme, which coordinated and promoted CMIP6 through its Working Group on Coupled Modelling. We thank the climate modeling groups for producing and making available their model output, the Earth System Grid Federation (ESGF) for archiving the data and providing access, and the multiple funding agencies who support CMIP6 and ESGF. The CMIP6 model data used in the paper is available from \href{https://esgf-node.llnl.gov/projects/cmip6/}{https://esgf-node.llnl.gov/projects/cmip6/} \citep{EyringEtAl2016}. We thank Urs Beyerle and Lukas Brunner for the preparation and maintenance of CMIP6 data. We acknowledge funding received from the Swiss Data Science Centre within the project ``Data Science-informed attribution of changes in the Hydrological cycle'' (DASH, C17-01). All methods and data needed to reproduce the results are described in the paper. The code to reproduce results and all figures are available at \href{https://github.com/eszekely/robustDA}{https://github.com/eszekely/robustDA}.}

\appendix
\renewcommand*\appendixpagename{\Large Appendices}
\renewcommand{\thesection}{\appendixname\ \Alph{section}}
\appendixpage
\section{Hypothesis testing procedure}
\label{sect:HT_appendix}

The entire set $\mathcal{M}$ of models available is split into two subsets, one subset $\mathcal{T}$ used for training and one subset $\mathcal{U}$ used for the hypothesis testing and prediction, $\mathcal{M} = \mathcal{T} \cup \mathcal{U}$. 

Let $t$ denote a test statistic drawn from a random variable $T$. We denote by $p_{m,0}$ the density of $t$ for a given climate model $m$ under the null hypothesis $H_0$. The expectation $\mu_{m,0}$ and variance $\sigma_{m,0}^2$ of the null distribution for model $m$ are
\begin{equation*}
    \begin{gathered}
        \mu_{m,0} = \mathbb{E}_{m,0}[T],\\
        \sigma_{m,0}^2 = \mathbb{E}_{m,0}[(T-\mu_{m,0})^2].    
    \end{gathered}
\end{equation*}

The average mean and variance of the null distribution for the set of models $m \in \mathcal{U}$ are
\begin{equation*}
    \begin{gathered}
        \mu_0 = \mathbb{E}_0[T] = \frac{1}{|\mathcal{U}|} \sum_{m \in \mathcal{U}} \mu_{m,0}, \\
        \sigma_0^2 = \mathbb{E}_0[(T-\mu_0)^2] = \frac{1}{|\mathcal{U}|} \sum_{m \in \mathcal{U}} \sigma_{m,0}^2 + \frac{1}{|\mathcal{U}|} \sum_{m \in \mathcal{U}} (\mu_{m,0} - \mu_0)^2.
    \end{gathered}
\end{equation*}

Let $\hat{\mu}_0$ and $\hat{\sigma}_0$ be estimators for $\mu_0$ and $\sigma_0$. We define the threshold $\theta$ for the hypothesis test as
\begin{equation}
\label{eq:ap_theta}
    \theta = \hat{\mu}_0 \pm z_{\alpha_*} \hat{\sigma}_0,
\end{equation}
where $z_{\alpha_*}$ is the quantile of level $1-\alpha_*$ of a standard normal Gaussian.

Given multiple runs per model, let $t_{m,r,0}$ and $t_{m,r,1}$ be the observed values of $t$ for model $m$ and run $r$ under the null and the alternative hypotheses, respectively. The maximum likelihood estimator of the mean is given by
\begin{equation*}
    \hat{\mu}_0 = \frac{1}{|\mathcal{U}|} \sum_{m \in \mathcal{U}} \hat{\mu}_{m,0}, \quad \text{with} \quad \hat{\mu}_{m,0} = \frac{1}{R_{m,0}} \sum_{r=1}^{R_{m,0}} t_{m,r,0},
\end{equation*}
where $R_{m,0}$ is the number of runs used for the null distribution, \eg control runs, for model $m$. The estimator for the variance is given by
\begin{equation*}
    \hat{\sigma}_0^2 = \frac{1}{|\mathcal{U}|} \sum_{m \in \mathcal{U}} \hat{\sigma}_{m,0}^2 + \frac{1}{|\mathcal{U}|} \sum_{m \in \mathcal{U}} (\hat{\mu}_{m,0} - \hat{\mu}_0)^2, \quad \text{with} \quad \hat{\sigma}_{m,0}^2 = \frac{1}{R_{m,0}} \sum_{r=1}^{R_{m,0}} (t_{m,r,0} - \hat{\mu}_{m,0})^2.
\end{equation*}
If only one control run is available, \ie $R_{m,0}=1$, then $\hat{\mu}_{m,0} = t_{m,1,0}$ and $\hat{\sigma}_{m,0}^2 = 0$.

We next estimate the type I error $\alpha_m$ and the power $\kappa_m$ for each model $m \in \mathcal{U}$ as
\begin{equation*}
\begin{split}
    \alpha_m & = P_{m,0}(T > \theta) = \mathbb{E}_{m,0} [\mathbbm{1}\{t > \theta\}] = \int_{\theta}^{\infty} p_{m,0}(t)dt, \\
    \kappa_m & = P_{m,1}(T > \theta) = \mathbb{E}_{m,1} [\mathbbm{1}\{t > \theta\}] = \int_{\theta}^{\infty} p_{m,1}(t)dt,
\end{split}
\end{equation*}
where $p_{m,1}$ is the distribution of $t$ for model $m$ under the alternative hypothesis, \ie forced run with the forcing of interest. We use a ``leave-one-model-out'' strategy and construct the critical region of the null distribution using all the models in $\mathcal{U}$ (none of them seen during training) except the model $m$ left out for testing, \ie the set of models $\mathcal{U} \setminus \{m\}$. The computation of the threshold $\theta$ in \eqref{eq:ap_theta} relies on the set of models $\mathcal{U} \setminus \{m\}$, and therefore we further denote it by $\theta_{-m}$. Similarly, the estimators $\hat{\mu}_0$ and $\hat{\sigma}_0$ are computed leaving out model $m$, \ie all runs associated with $m$. The ``leave-one-model-out'' strategy avoids any overlaps between the construction of the critical region and the estimation of the type I error and power. This strategy is straightforward to apply to observations if one considers the observations as the model left out.

Let $z_{m,r,0} = \mathbbm{1} \{ t_{m,r,0} > \theta_{-m}\}$ and $z_{m,r,1} = \mathbbm{1} \{ t_{m,r,1} > \theta_{-m}\}$ be the binary variables indicating a rejection for a control run $r$ and a forced run $r$ of model $m$, respectively. The estimators of the type I error and power for each model $m$ are
\begin{equation*}
\begin{split}
    \hat{\alpha}_m &= \frac{1}{R_{m,0}} \sum_{r=1}^{R_{m,0}} z_{m,r,0}, \\
    \hat{\kappa}_m &= \frac{1}{R_{m,1}} \sum_{r=1}^{R_{m,1}} z_{m,r,1},
\end{split}
\end{equation*}
where $R_{m,1}$ is the number of forced runs for model $m$. To avoid estimation issues when $R_{m,0}$ and  $R_{m,1}$ are too low (\eg if only one control run or only one forced run per model are available), we repeat the procedure above $B$ times by constructing $B$ subsamples of models of size $|\mathcal{T}| = |\mathcal{M}| - |\mathcal{U}|$, drawn randomly without replacement from the entire set of models $\mathcal{M}$. The estimators of the type I error and power for each model computed from the $B$ different subsamples are then averaged using an aggregating method called subagging \citep[subsample aggregating,][]{BuehlmannYu2002}. We use here half-subagging where the size of the training set is $|\mathcal{T}|=\lceil |\mathcal{M}|/2 \rceil$ (using the ceiling function which rounds up to the closest integer) such that half of the models is used for training to learn the fingerprint and the other half of the models is used for the hypothesis testing. We obtain for each $m \in \mathcal{M}$:
\begin{equation*}
\begin{split}
    \hat{\alpha}_m = \frac{1}{B} \frac{1}{R_{m,0}} \sum_{b=1}^B \sum_{r=1}^{R_{m,0}} z_{m,r,0}^b, \quad &\text{with} \quad z_{m,r,0}^b = \mathbbm{1} \{ t_{m,r,0}^b > \theta_{-m}^b\},\\
    \hat{\kappa}_m = \frac{1}{B} \frac{1}{R_{m,1}} \sum_{b=1}^B \sum_{r=1}^{R_{m,1}} z_{m,r,1}^b, \quad &\text{with} \quad z_{m,r,1}^b = \mathbbm{1} \{ t_{m,r,1}^b > \theta_{-m}^b\}.
\end{split}
\end{equation*}

The average type I error and power across all the models in $\mathcal{M}$ are
\begin{equation*}
    \begin{split}
        \hat{\alpha} &= \frac{1}{|\mathcal{M}|} \sum_{m \in \mathcal{M}} \hat{\alpha}_m, \\
        \hat{\kappa} &= \frac{1}{|\mathcal{M}|} \sum_{m \in \mathcal{M}} \hat{\kappa}_m.
    \end{split}
\end{equation*}

\section{Anchor regression and mean independence}
\label{sect:ap_nonlinearAnchors}

The orthogonal projection of the residuals $R$ on the subspace of every possible function of $A$ is the conditional expectation $P_A(R) = \mathbb{E}[R \mid A]$. Using the projection $P_A$, the population version of anchor regression from \eqref{eq:anchorPA} can be rewritten as
\begin{equation}
\label{eq:anchorPA_condExp}
\begin{split}
\beta^{\gamma} & = \operatorname*{argmin}_{\beta} \mathbb{E}[((I - P_A)(R))^2] + 
					\gamma \mathbb{E}[(P_A(R))^2]\\
     & = \operatorname*{argmin}_{\beta} \mathbb{E}[(R - \mathbb{E}[R \mid A])^2] + 
					\gamma \mathbb{E}[\mathbb{E}^2[R \mid A]].\\
\end{split}
\end{equation}

Using the law of total expectation and the fact that the expected value of the residuals is $\mathbb{E}[R] = 0$, the variance of the conditional expectations $\mathbb{E}[R \mid A]$ can be written as
\begin{equation}
\label{eq:varDecomp}
\begin{split}
    \text{Var}(\mathbb{E}[R \mid A]) & = \mathbb{E}[\mathbb{E}^2[R \mid A]] - \mathbb{E}^2[\mathbb{E}[R \mid A]]\\
                        & = \mathbb{E}[\mathbb{E}^2[R \mid A]] - \mathbb{E}^2[R]\\
                        & = \mathbb{E}[\mathbb{E}^2[R \mid A]],
\end{split}
\end{equation}
where the inner expectation is the conditional expectation of the residuals given the anchor $\mathbb{E}[R \mid A]$. The penalty term in \eqref{eq:anchorPA_condExp} is thus equivalent to the variance of the conditional expectations, and replacing \eqref{eq:varDecomp} into \eqref{eq:anchorPA_condExp} we obtain the anchor regression formulation from \eqref{eq:anchorPA}. As discussed in Sect.~\ref{subsect:nonlinearAnchor}, minimizing the variance of the conditional expectations leads to residuals that are mean independent of the anchor. 

One measure of mean independence is the \textit{correlation ratio} $\eta_A(R)$ defined by \citet{Renyi1959} as
\begin{equation}
\label{eq:corrRatio}
    \eta_A(R) = \eta(R \mid A) = \sqrt{\frac{\text{Var}(\mathbb{E}[R \mid A])}{\text{Var}(R)}},
\end{equation}
which is zero if and only if $R$ is mean independent of $A$, \ie its conditional mean equals its (unconditional) mean. Both mean independence and correlation ratio are asymmetric in nature, while independence and uncorrelatedness are both symmetric. Mean independence being an asymmetric property, the fact that the residuals are mean independent of the anchor does not imply that the converse is true, \ie the anchor is not mean independent of the residuals. Often, minimizing $\text{Var}(\mathbb{E}[R \mid A])$ in anchor regression with increasing $\gamma$ leads to an increase in the total variance of the residuals, $\text{Var}(R)$, due to the trade-off between robustness and accuracy. Therefore, minimizing $\text{Var}(\mathbb{E}[R \mid A])$ often leads to minimizing the ratio above, that is, the correlation ratio.

\citet{Renyi1959} showed that the correlation ratio from \eqref{eq:corrRatio} can be written as the supremum over the linear correlations between $R$ and all functions $h(A)$,
\begin{equation}
\label{eq:corrRatio_sup}
    \eta(R \mid A) = \sup_h \rho(R, h(A)),
\end{equation}
for all $h(A)$ with finite positive variance. Minimizing the projection of $R$ on each of these functions often leads to an increase in the angle between the component of the residuals that projects on $h(A)$ and the total residuals, and therefore to a decrease of the cosine between the two vectors. This is due to the fact that the residuals tend to increase faster than their projection on $h(A)$ when the constraint to reduce the projection is strongly enforced, \ie large $\gamma$ in anchor regression. Because $R$ is already centered, if we additionally center $h(A)$, the cosine is equal to the correlation $\rho(R, h(A))$. Therefore, the objective of anchor regression is tightly related to reducing the correlation of $R$ with the different functions $h(A)$. Using \eqref{eq:corrRatio_sup}, this in turn helps to reduce the correlation ratio and therefore make $R$ mean independent of $A$ when the correlation ratio becomes zero. 

For a zero mean and unit variance function $h(A)$, the coefficient of the component of $R$ along $h(A)$ is $c = \mathbb{E}[Rh(A)]$, and the projection of $R$ on $h(A)$ is of the form $ch(A)$. If all these coefficients are zero and given that the mean of the residuals is zero, $\mathbb{E}[R] = 0$, then we have $\mathbb{E}[R \mid A] = 0$, thus achieving mean independence. If $R$ is mean indepedent of $A$, we additionally have 
\begin{equation*}
    \mathbb{E}[Rh(A)] = \mathbb{E}[R]\mathbb{E}[h(A)],
\end{equation*}
meaning that $R$ and $h(A)$ are uncorrelated. Among all functions $h(A)$, the function that maximizes the correlation with $A$ is the conditional expectation,
$$\sup_h \rho(R, h(A)) = \rho(R, \mathbb{E}[R \mid A]).$$

From an implementation perspective, the procedure to minimize the correlation ratio is described in Sect.~\ref{subsect:nonlinearAnchor}, where we add to the linear term $A$ additional nonlinear terms through the functions $h(A)$ to obtain the anchor matrix $\A_h$. The more nonlinear terms we add, the closer the projection of the residuals onto the space spanned by these terms gets to the conditional expectations $\mathbb{E}[R \mid A]$. With increasing $\gamma$, all these projections are encouraged to be as small as possible. 

In the general case, the conditional expectation $\mathbb{E}[R \mid A]$ is a nonlinear function of $A$, but if $\mathbb{E}[R \mid A]$ is a linear function of $A$, \eg $(R,A)$ are bivariate Gaussian, 
$$\mathbb{E}[R \mid A] = a A + b,$$ 
the square of the correlation ratio is equal to the square of the correlation between $R$ and $A$,
\begin{equation}
\label{eq:ratioVar}
    \frac{\text{Var}(\mathbb{E}[R \mid A])}{\text{Var}(R)} = \rho^2(R,A).
\end{equation}
Thus, minimizing the correlation or the correlation ratio lead to the same results, \ie $R$ is mean independent of $A$ (and implicitly decorrelated). In our case, because $\mathbb{E}[R] = 0$ and since we center the anchor variable such that $\mathbb{E}[A] = 0$, the correlation between $R$ and $A$ is given by
$$\rho(R,A) = \frac{\text{Cov}(R,A)}{\sqrt{\text{Var}(R)} \sqrt{\text{Var}(A)}} 
            = \frac{\mathbb{E}[(R - \mathbb{E}[R])(A - \mathbb{E}[A])]}{\sqrt{\mathbb{E}[R^2]} \sqrt{\mathbb{E}[A^2]}}
            = \frac{\mathbb{E}[RA]}{\sqrt{\mathbb{E}[R^2]} \sqrt{\mathbb{E}[A^2]}}.$$
If $\mathbb{E}[RA]=0$ then $\rho(R,A) = 0$. In the case of the linear anchor, penalizing $P_A(R)$ is equivalent to penalizing $\mathbb{E}[RA]^2$ which minimizes the component of $R$ that is along $A$ and will tend to encourage that the residuals $R$ are decorrelated from $A$. 

Using the law of total variance or variance decomposition formula, we can also decompose the variance of the residuals as
\begin{equation}
\label{eq:lawVar}
\text{Var}(R) = \underbrace{\mathbb{E}[\text{Var}(R \mid A)]}_{\substack{\text{part of $R$} \\ \text{``unexplained'' by $A$}}} + \ \underbrace{\text{Var}(\mathbb{E}[R \mid A])}_{\substack{\text{part of $R$} \\ \text{``explained'' by $A$}}},    
\end{equation}
where the first term corresponds to the part of the residuals that are not explained by the anchor $A$, and the second term to the part of the residuals explained by the anchor. Thus, as $\gamma$ increases, by penalizing $\text{Var}(\mathbb{E}[R \mid A])$  more and more, anchor regression tries to reduce the part of the residuals explained by $A$, in addition to reducing the residuals through the $L_2$-loss function.

\section{Forcings and temperature correlation maps}
\label{sect:ap_forcing_corr}

\begin{figure}[ht]
    \centering
    \includegraphics[width = 0.55\textwidth]{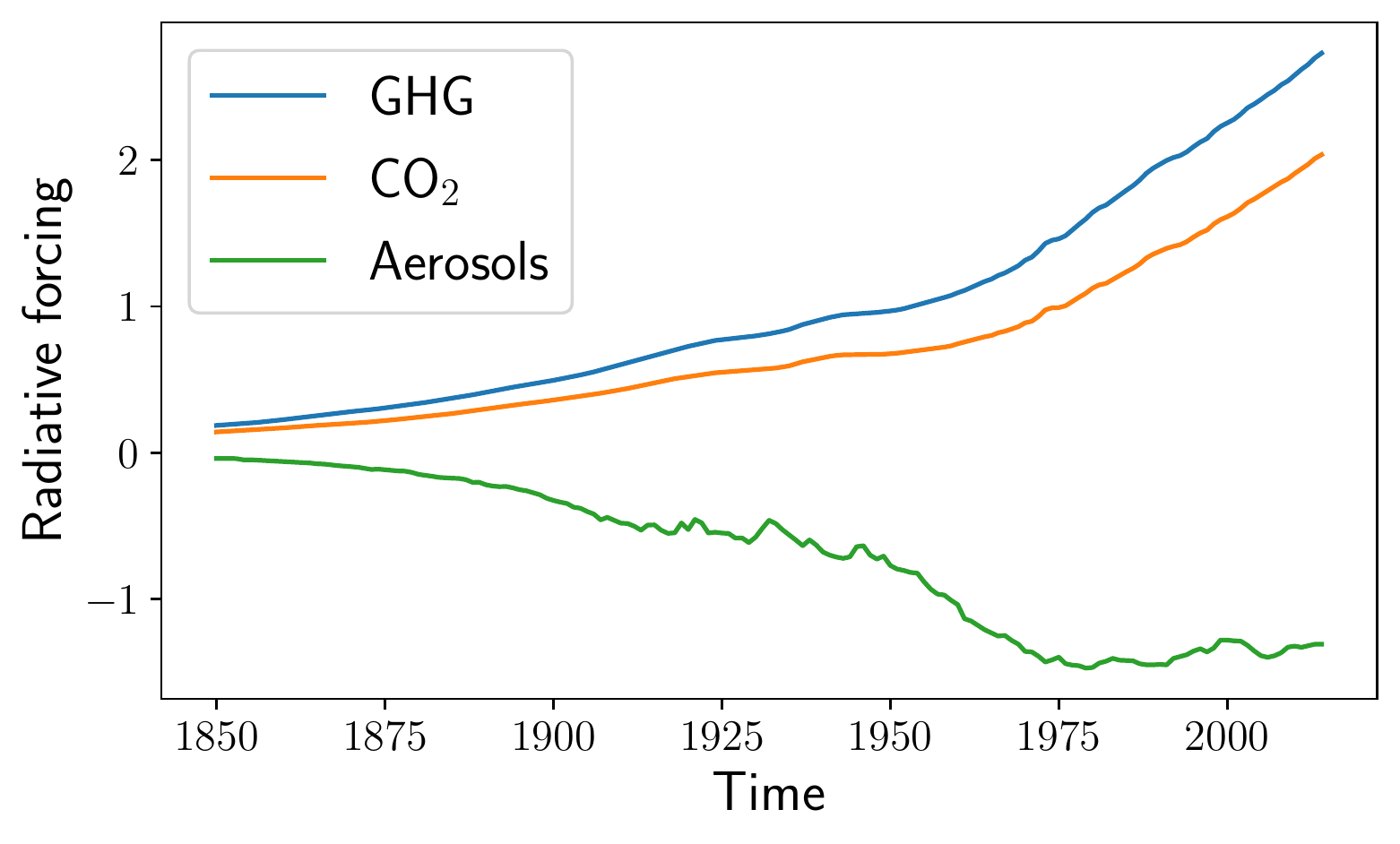}
    \caption{Radiative forcings used in the experiments (GHG, CO$_2$, aerosols).}
    \label{fig:radForcings}
\end{figure}

\begin{figure}[ht]
    \centering{
    \subfloat[GHG]{\includegraphics[width = 0.333\textwidth]{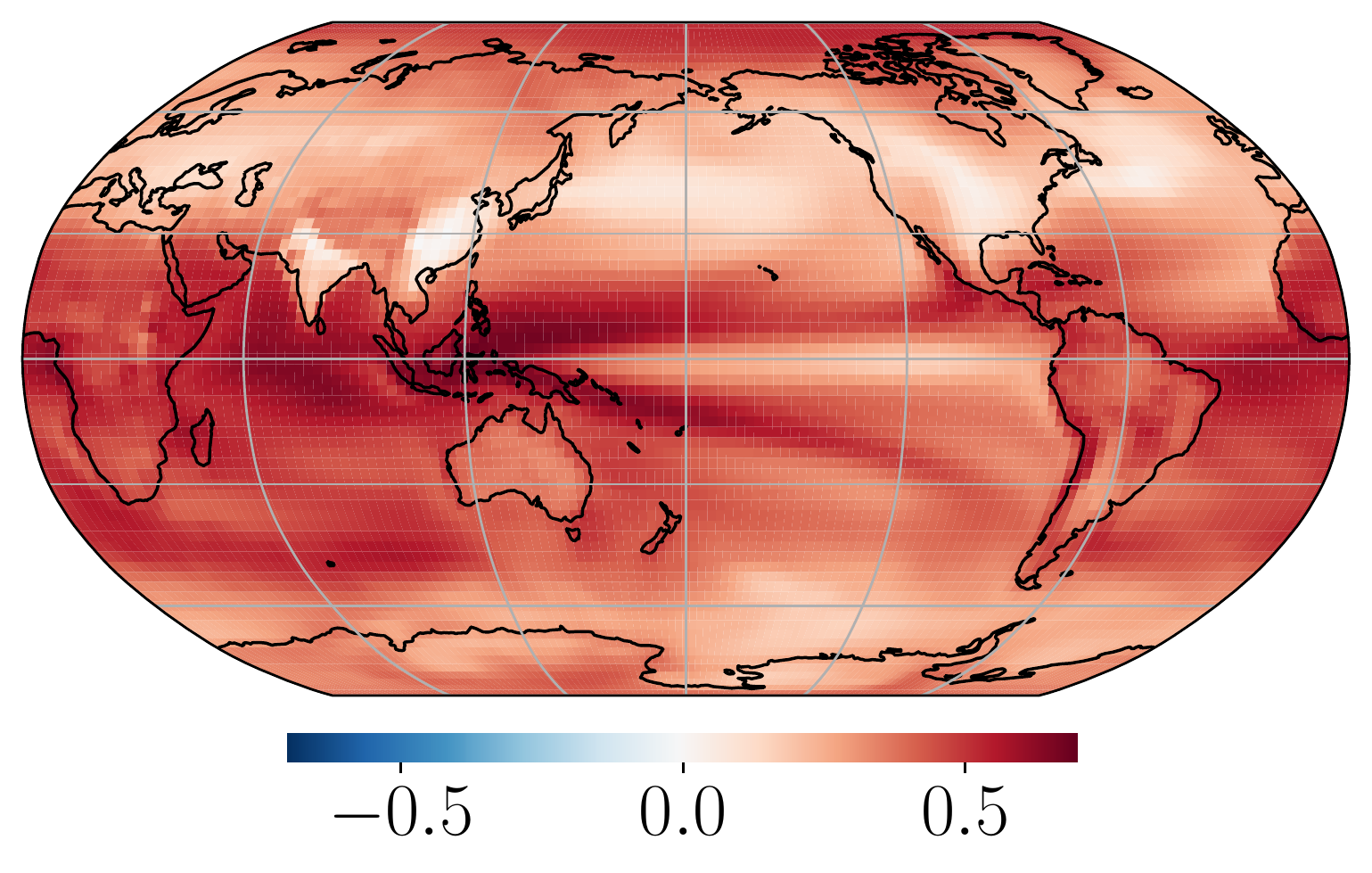}}
	\subfloat[CO$_2$]{\includegraphics[width = 0.333\textwidth]{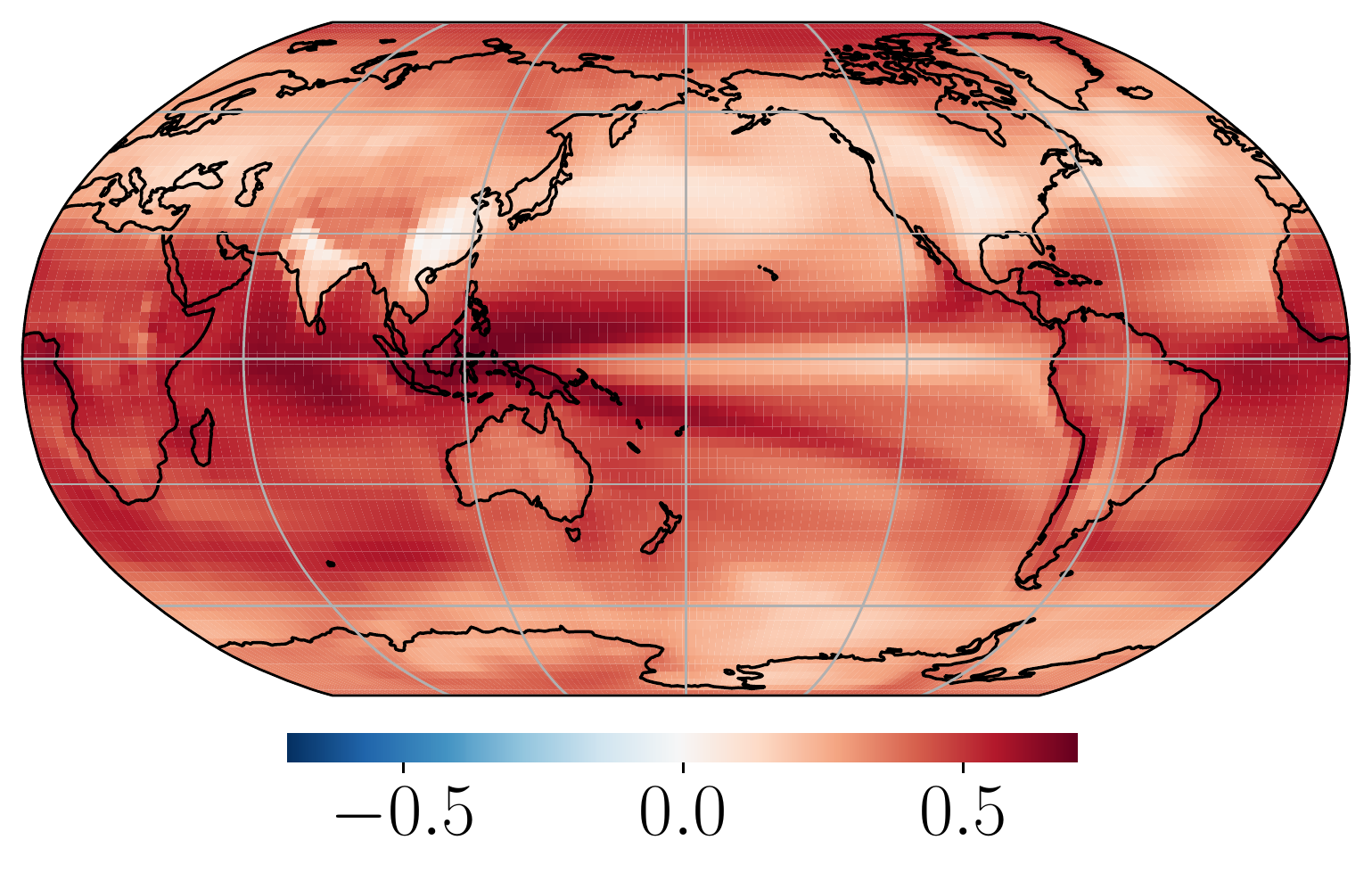}}
    \subfloat[Aerosols]{\includegraphics[width = 0.333\textwidth]{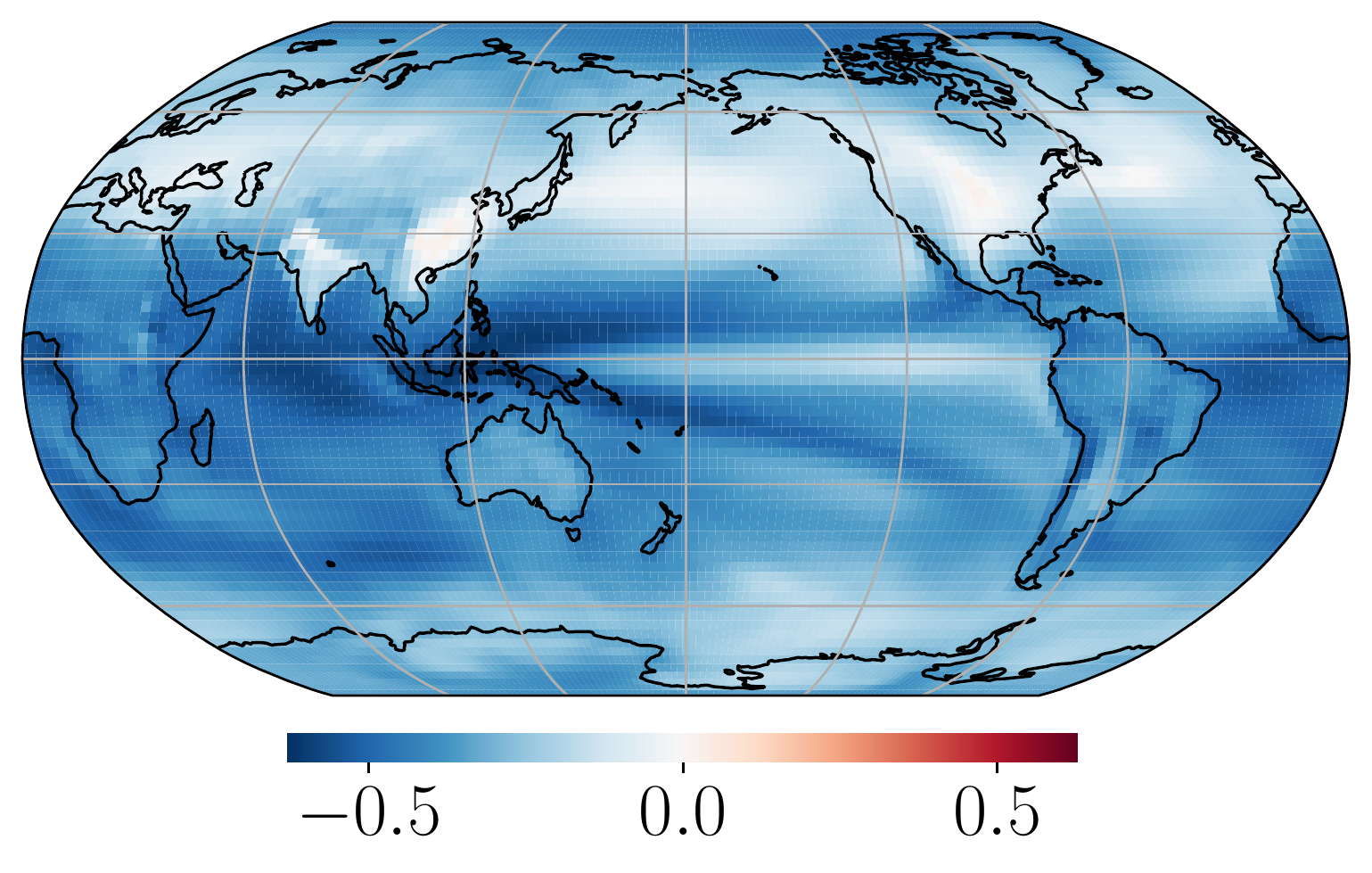}}}
    \caption{Correlation maps of temperature with different forcings (GHG, CO2, aerosols). 
	\label{fig:corr_forcings}}
\end{figure}

In this appendix we show the time series of the radiative forcings for the historical period from 1850--2014 (Fig.~\ref{fig:radForcings}), and the correlation maps of the temperature with these forcings (Fig.~\ref{fig:corr_forcings}).

\section{Models}
\label{sect:ap_models}

Table \ref{tab:models} shows the climate models and their variants used in the experimental section.

\begin{table}[ht]
    \centering
    \begin{tabular}{p{2.5cm}p{11cm}}
         Model & Model variants \\
         \hline
         ACC & ACCESS-CM2, ACCESS-ESM1-5\\
         AWI & AWI-CM-1-1-MR, AWI-ESM-1-1-LR\\
         BCC & BCC-CSM2-MR, BCC-ESM1 \\ 
         CAM & CAMS-CSM1-0 \\ 
         CAS & CAS-ESM2-0 \\ 
         CES & CESM2, CESM2-FV2, CESM2-WACCM, CESM2-WACCM-FV2 \\ 
         CIE & CIESM \\ 
         CMC & CMCC-CM2-HR4, CMCC-CM2-SR5, CMCC-ESM2 \\ 
         Can & CanESM5 \\ 
         E3S & E3SM-1-0, E3SM-1-1, E3SM-1-1-ECA \\ 
         EC- & EC-Earth3, EC-Earth3-AerChem, EC-Earth3-Veg, EC-Earth3-Veg-LR \\ 
         FGO & FGOALS-f3-L, FGOALS-g3 \\ 
         FIO & FIO-ESM-2-0 \\ 
         GFD & GFDL-CM4, GFDL-ESM4  \\ 
         GIS & GISS-E2-1-G, GISS-E2-1-G-CC, GISS-E2-1-H, GISS-E2-2-H \\ 
         IIT & IITM-ESM \\ 
         INM & INM-CM4-8, INM-CM5-0 \\ 
         IPS & IPSL-CM5A2-INCA, IPSL-CM6A-LR, IPSL-CM6A-LR-INCA \\ 
         KIO & KIOST-ESM  \\ 
         MCM & MCM-UA-1-0 \\ 
         MIR & MIROC6 \\ 
         MPI & MPI-ESM-1-2-HAM, MPI-ESM1-2-HR, MPI-ESM1-2-LR  \\ 
         MRI & MRI-ESM2-0 \\ 
         NES & NESM3 \\
         Nor & NorCPM1, Nor-ESM1-F, Nor-ESM2-LM, Nor-ESM2-MM \\
         SAM & SAM0-UNICON \\
         Tai & TaiESM1 \\
    \end{tabular}
    \caption{Models and their variants.}
\label{tab:models}
\end{table}

\section{Additional results}
\label{sect:ap_addResults}

\begin{figure}
    \includegraphics[width = \textwidth]{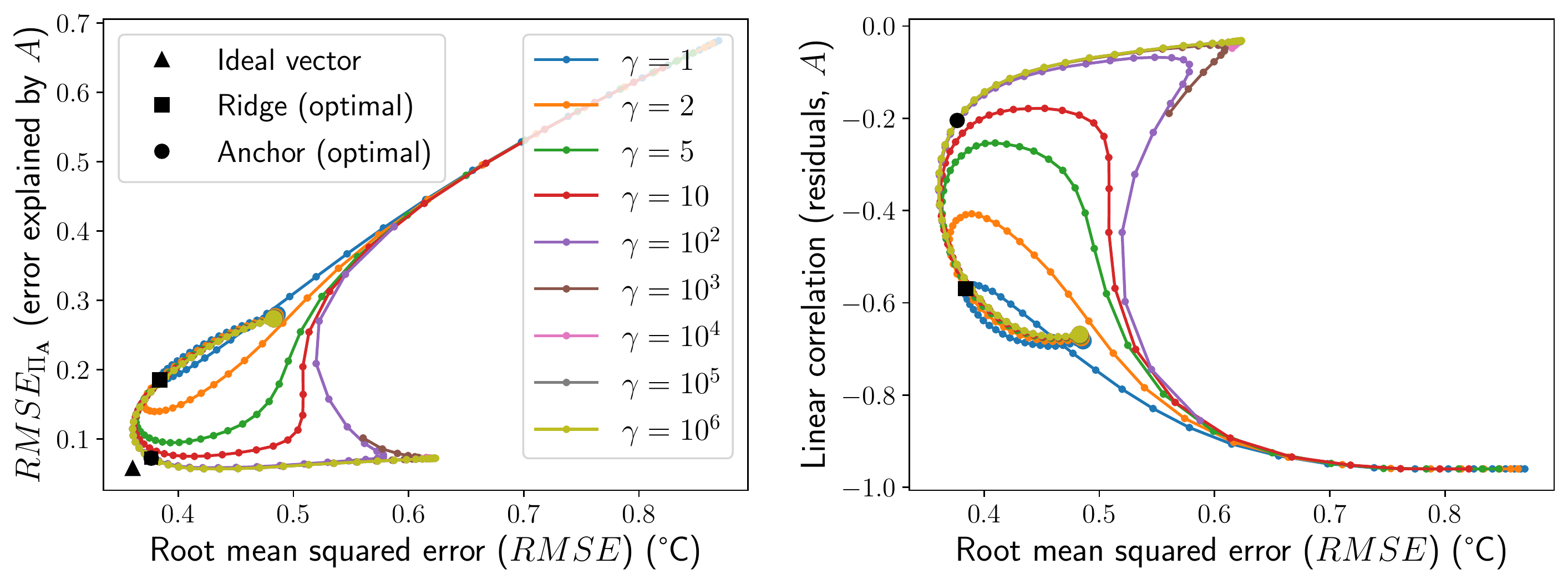}
	\caption{Predictions on the test data for the GHG forcing with the aerosols linear anchor. 
	\label{fig:GHG_aer_lin_test}}
\end{figure}

\begin{figure}
    \includegraphics[width = 0.95\textwidth]{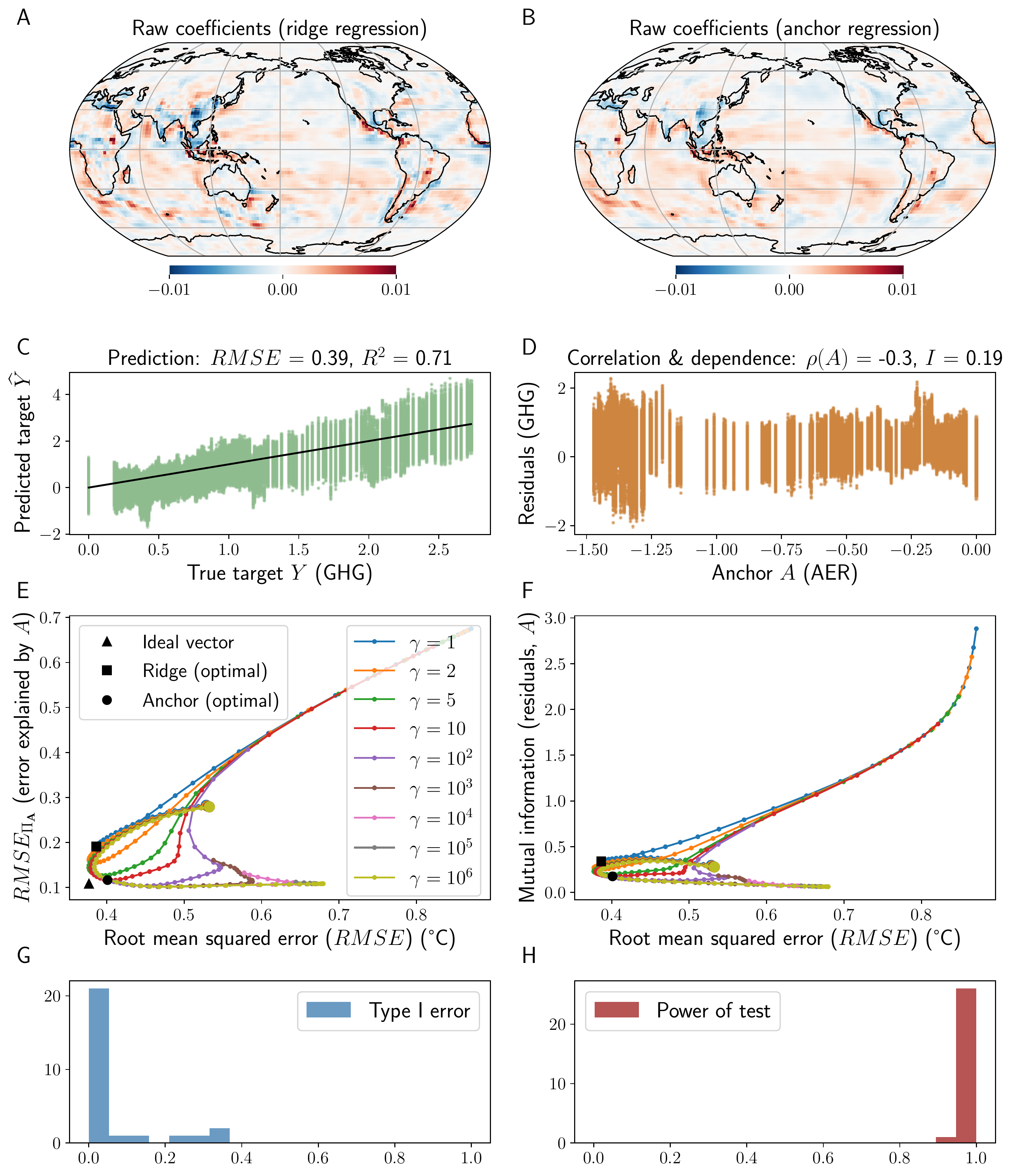}
	\caption{Prediction and attribution of the GHG forcing with aerosols nonlinear (square) anchors. Panels similar to Fig.~\ref{fig:GHG_aer_lin}.
	\label{fig:GHG_aer_nonlin}}
\end{figure}

\begin{figure}
    \includegraphics[width = \textwidth]{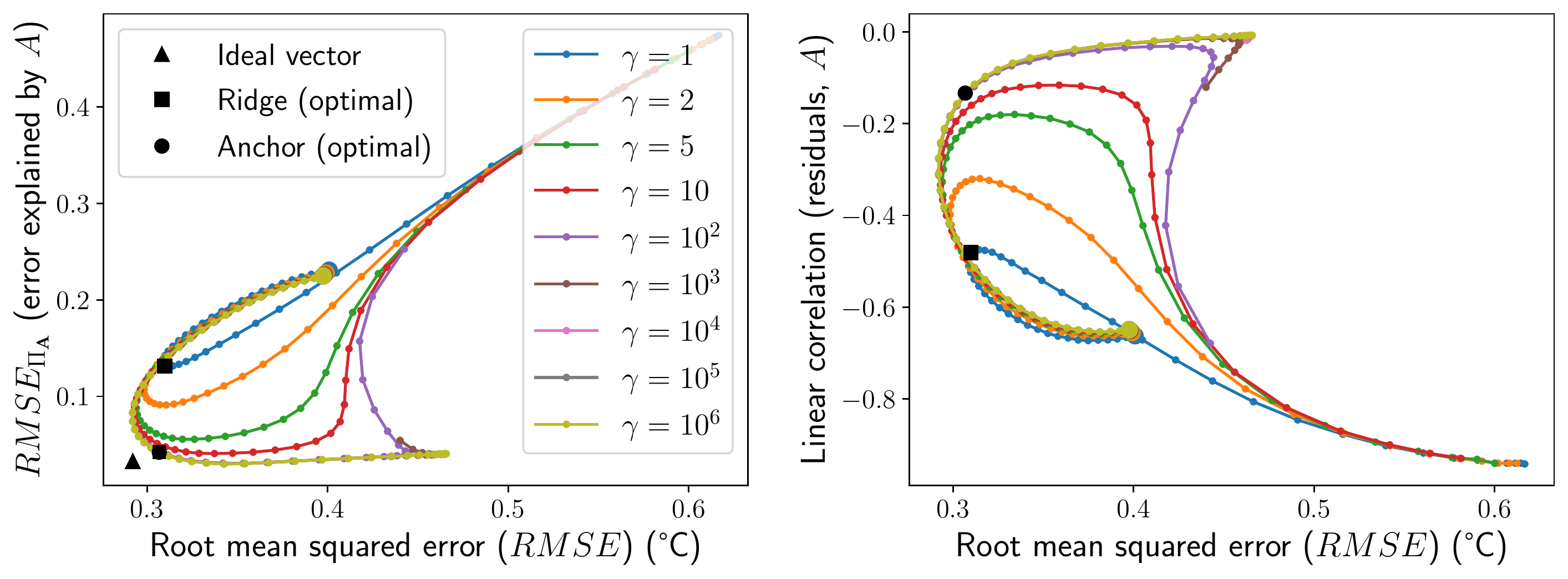}
	\caption{Predictions on the test data for the aerosols forcing with the linear CO$_2$ anchor. 
	\label{fig:aer_co2_lin_test}}
\end{figure}

\begin{figure}
    \includegraphics[width = \textwidth]{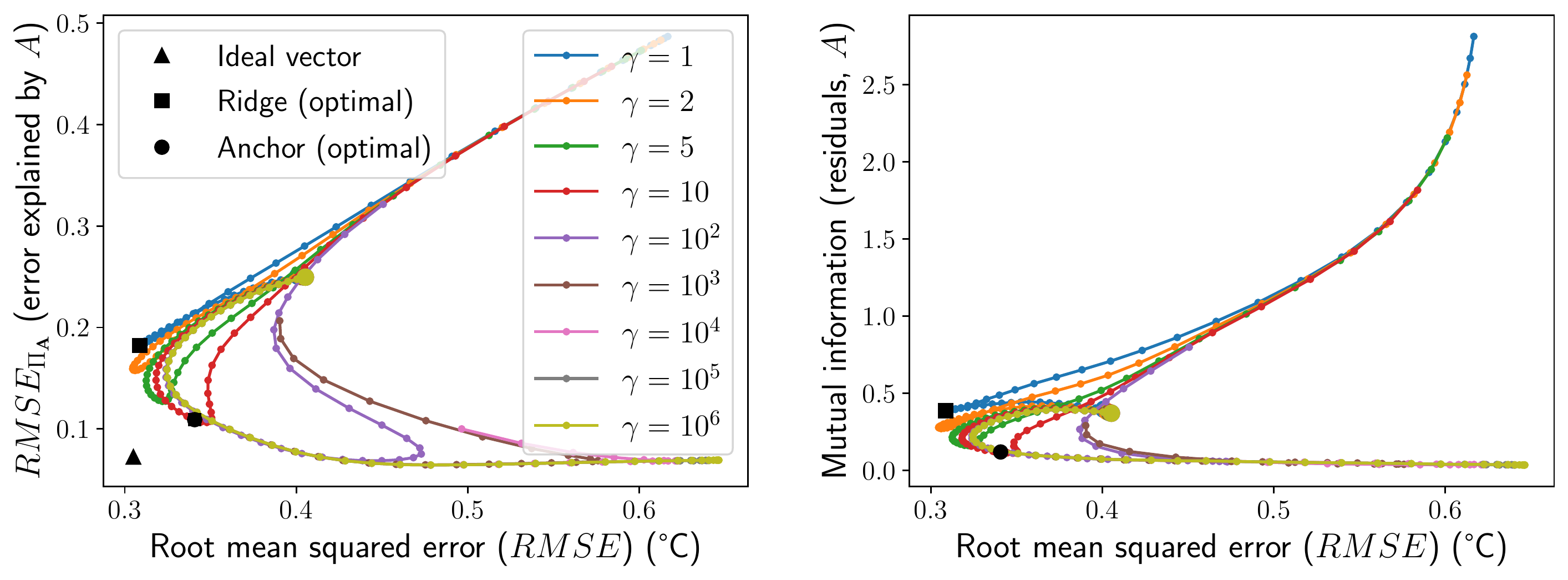}
	\caption{Predictions on the test data for the aerosols forcing with the nonlinear CO$_2$ anchors. 
	\label{fig:aer_co2_nonlin_test}}
\end{figure}

Figures \ref{fig:GHG_aer_lin_test}, \ref{fig:GHG_aer_nonlin}, \ref{fig:aer_co2_lin_test} and \ref{fig:aer_co2_nonlin_test} show additional results for the attribution of the GHG and aerosols forcings.

\vskip 0.2in
\bibliography{bibliography}

\end{document}